\documentclass[times,twocolumn,final]{elsarticle}

\usepackage{medima}
\usepackage{framed,multirow}

\usepackage{latexsym}

\usepackage{xcolor}

\usepackage{hyperref}

\usepackage{amsmath,amssymb,amsfonts}
\usepackage{algorithmic}
\usepackage{graphicx}
\usepackage{textcomp}
\usepackage[textsize=tiny]{todonotes}
\usepackage{threeparttable}
\usepackage{caption}
\usepackage{subcaption}
\usepackage{adjustbox}
\usepackage{float}
\usepackage{url}
 \usepackage{comment}
\makeatletter
\def\UrlAlphabet{%
      \do\a\do\b\do\c\do\d\do\e\do\f\do\g\do\h\do\i\do\j%
      \do\k\do\l\do\m\do\n\do\o\do\p\do\q\do\r\do\s\do\t%
      \do\u\do\v\do\w\do\x\do\y\do\z\do\A\do\B\do\C\do\D%
      \do\E\do\F\do\G\do\H\do\I\do\J\do\K\do\L\do\M\do\N%
      \do\O\do\P\do\Q\do\R\do\S\do\T\do\U\do\V\do\W\do\X%
      \do\Y\do\Z}
\def\UrlDigits{\do\1\do\2\do\3\do\4\do\5\do\6\do\7\do\8\do\9\do\0}
\g@addto@macro{\UrlBreaks}{\UrlOrds}
\g@addto@macro{\UrlBreaks}{\UrlAlphabet}
\g@addto@macro{\UrlBreaks}{\UrlDigits}
\makeatother

\definecolor{newcolor}{rgb}{.8,.349,.1}

\journal{Medical Image Analysis}

\begin{document}

\verso{Haomiao Ni \textit{et~al.}}

\begin{frontmatter}


\title{Semi-supervised Body Parsing and Pose Estimation for Enhancing Infant General Movement Assessment}

\author[1]{Haomiao \snm{Ni}\fnref{fn1}}
\author[1,2]{Yuan \snm{Xue}\fnref{fn1}}
\fntext[fn1]{These authors contributed equally to this work.}
\author[3]{Liya \snm{Ma}}
\author[4]{Qian \snm{Zhang}}
\author[3]{Xiaoye \snm{Li}\corref{cor1}}
\author[1]{Xiaolei \snm{Huang}\corref{cor1}
\cortext[cor1]{Corresponding author: Xiaoye Li (xiaoyelisz@foxmail.com) and Xiaolei Huang (suh972@psu.edu)}}

\address[1]{College of Information Sciences and Technology, The Pennsylvania State University, University Park, PA, USA}
\address[2]{Department of Electrical and Computer Engineering, Johns Hopkins University, Baltimore, MD, USA}
\address[3]{Shenzhen Baoan Women’s and Children’s Hospital, Jinan University, Shenzhen, China}
\address[4]{School of Information and Control Engineering, Xi’an University of Architecture and Technology, Xi’an, China}

\received{}
\finalform{}
\accepted{}
\availableonline{}
\communicated{}

\begin{abstract}
General movement assessment (GMA) of infant movement videos (IMVs) is an effective method for early detection of cerebral palsy (CP) in infants.  We demonstrate in this paper that end-to-end trainable neural networks for image sequence recognition can be applied to achieve good results in GMA, and more importantly, augmenting raw video with infant body parsing and pose estimation information can significantly improve performance. To solve the problem of efficiently utilizing partially labeled IMVs for body parsing, we propose a semi-supervised model, termed \textit{SiamParseNet} (SPN), which consists of two branches, one for intra-frame body parts segmentation and another for inter-frame label propagation. During training, the two branches are jointly trained by alternating between using input pairs of only labeled frames and input of both labeled and unlabeled frames. We also investigate training data augmentation by proposing a factorized video generative adversarial network (FVGAN) to synthesize novel labeled frames for training. FVGAN decouples foreground and background generation which allows for generating multiple labeled frames from one real labeled frame. When testing, we employ a multi-source inference mechanism, where the final result for a test frame is either obtained via the segmentation branch or via propagation from a nearby \textit{key} frame. We conduct extensive experiments for body parsing using SPN on two infant movement video datasets; on these partially labeled IMVs, we show that SPN coupled with FVGAN achieves state-of-the-art performance. We further demonstrate that our proposed SPN can be easily adapted to the infant pose estimation task with superior performance. Last but not least, we explore the clinical application of our method for GMA. We collected a new clinical IMV dataset with GMA annotations, and our experiments show that our SPN models for body parsing and pose estimation trained on the first two datasets generalize well to the new clinical dataset and their results
can significantly boost the convolutional recurrent neural network (CRNN) based {GMA} prediction performance when combined with raw video inputs.
\end{abstract}



\end{frontmatter}


\section{Introduction}

\begin{figure*}[t]
    \centering
    \includegraphics[width=0.80\linewidth]{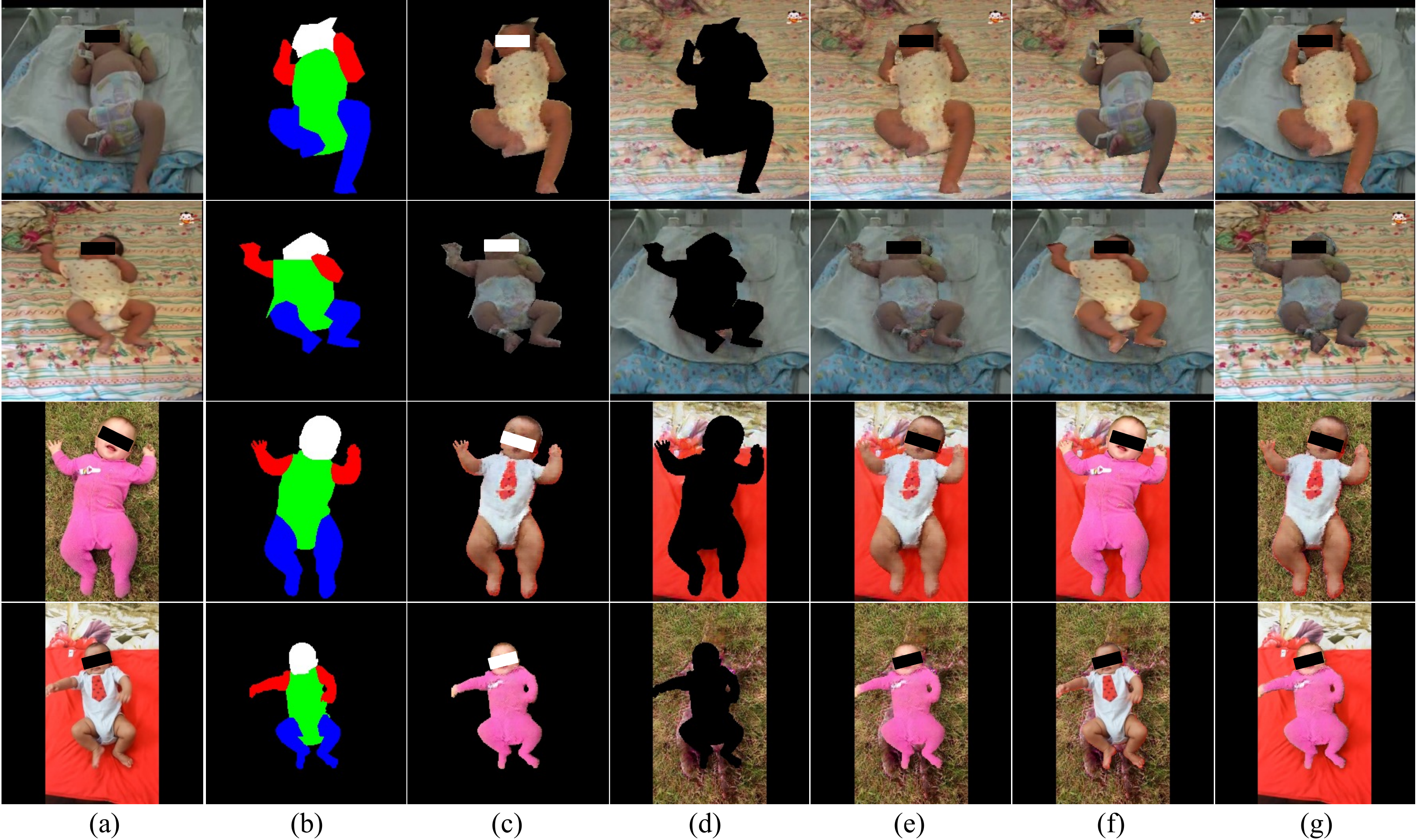}
    \caption{{
     Example synthesized results of FVGAN on the BHT dataset (1st and 2nd rows) and the Youtube-Infant dataset (3rd and 4th rows).} (a) Original images; (b) Ground truth masks; (c) Synthesized foreground;
    (d) Synthesized background;
    (e) Synthesized foreground plus synthesized background $I_{\text{whole}}$;
   (f) Synthesized background plus original foreground $I_{\text{back}}$;
    (g) Synthesized foreground plus original background $I_{\text{fore}}$. Note that FVGAN can generate 3 new labeled frames (e-g) using just one existing labeled frame (a).}
    \label{fig:fvgan_res2}
\end{figure*}

\begin{figure*}[t]
    \centering
    \includegraphics[width=0.85\linewidth]{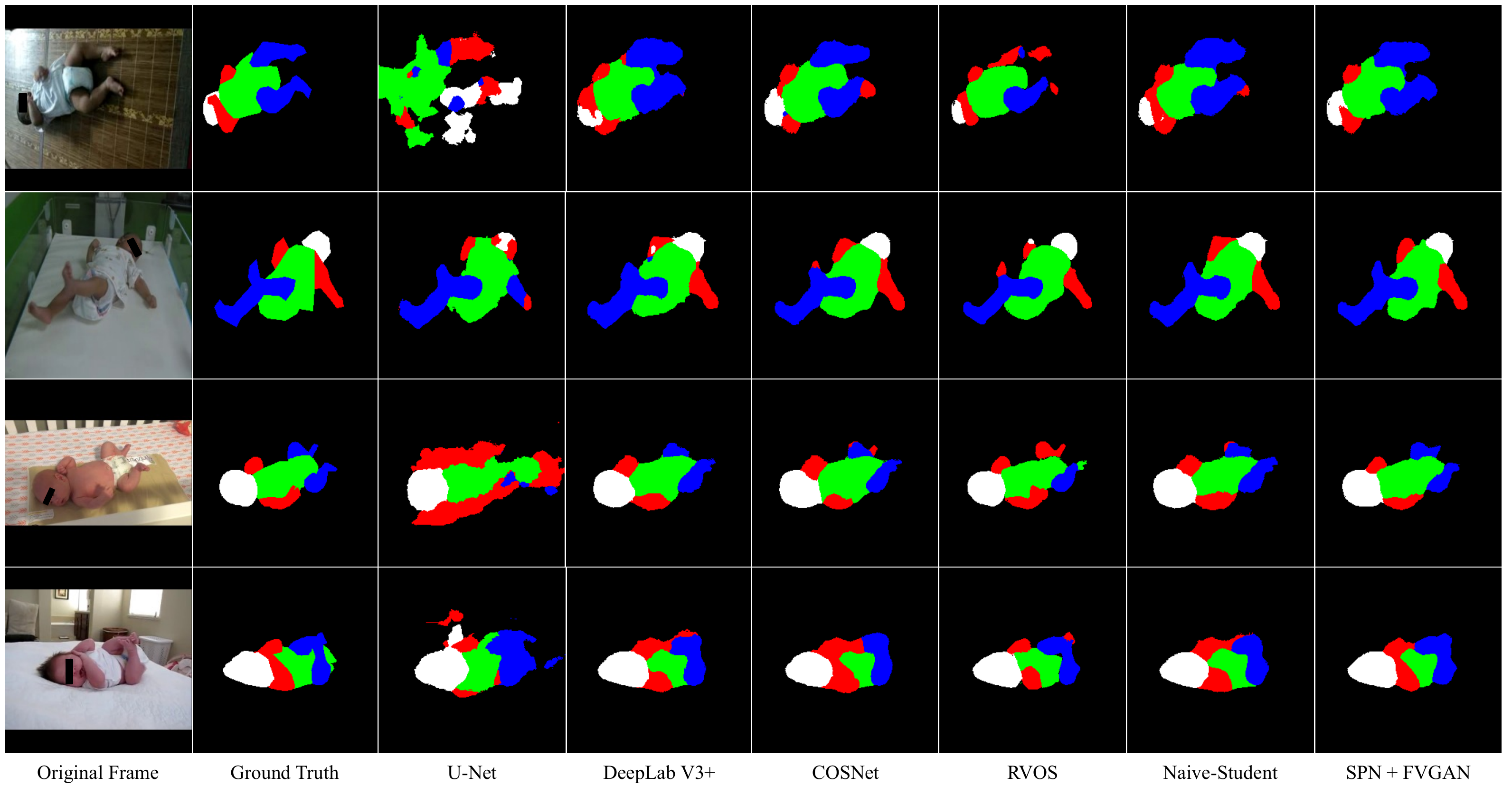}
    \caption{
    {
    Comparison between different body parsing methods on BHT dataset (1st and 2nd row) and Youtube-Infant (3rd and 4th row).
    }
    }
    \label{fig:sota}
\end{figure*}

\label{sec:introduction}
Cerebral palsy (CP) is a group of neurodevelopmental disorders that appear in early childhood, which cause limitations to body movement and motion coordination \citep{richards2013cerebral}.
Early detection of CP enables early rehabilitative interventions for high-risk infants. One effective CP detection approach is general movement assessment (GMA) \citep{prechtl1990qualitative,adde2007general}, where clinicians observe the movement of infant body parts in infant movement videos (IMVs) and evaluate the corresponding CP scores. {Such a manual assessment task is tedious to clinicians and can be subjective and error-prone. Automated or computer-assisted
GMA can potentially alleviate the burden for clinicians and make the process more efficient and objective.}

{The input to an automated or computer-assisted GMA system can be raw videos, or raw videos combined with various extracted semantic representations, such as infant body part segmentation map (\textit{i.e.} body parsing results) and body pose (\textit{i.e.} pose estimation results).
Inferring body part segmentation or pose from IMVs is a non-trivial task however.}
With body parsing, one common challenge is the frequent occlusions among infant body parts due to their spontaneous movements, which makes it difficult to model the temporal correspondence between video frames. Moreover, towards training an automated body parsing system for IMVs, the annotation of segmented body parts in IMVs is expensive to acquire \citep{zhang2019online}.

 
{In this paper, we aim to develop a robust and general framework for both infant body parsing and pose estimation and further
evaluate the benefit of adding body parsing and pose estimation information to enhance the performance of neural-network based automatic GMA. This work is significantly different from our prior work~\citep{Ni2020SiamParseNetJB}; while the previous work introduces our proposed \textit{SiamParseNet} (SPN) for infant body parsing and demonstrates its segmentation performance on one small IMV dataset, in this work we propose an effective Factorized Video Generative Adversarial Network (FVGAN) for augmented training, demonstrate how to adapt SPN to the task of pose estimation, train and test SPN models for body parsing and pose estimation tasks using two IMV datasets. Since GMA has been used as a standard method for early prediction of CP in infants with high risk of developing neurological dysfunctions~\citep{adde2007general,stoen2019predictive}, we further collect an IMV dataset containing 161 infant videos with GMA annotations to validate our proposed methods. Experiments on the clinical dataset show that the trained SPN models can generalize well to new clinical data and their results can significantly improve {GMA} prediction performance.}

More specifically, the contributions of this work include:
\begin{itemize}
    \item {We propose for the first time a semi-supervised learning framework for robust and accurate infant body parsing from IMVs, demonstrate its applicability to infant pose estimation, and show that combining body parsing and pose estimation results with raw videos leads to superior performance for GMA and CP risk prediction.} 
    \item {We conduct extensive experiments to evaluate our proposed body parsing and pose estimation framework and its application to GMA using three datasets from diverse sources. Our framework is shown to have good generalizability and models trained using two existing datasets in a semi-supervised manner can be directly applied to a new clinical dataset without fine-tuning and produce high-quality results. }
    \item Our proposed infant body parsing framework, termed SiamParseNet (SPN), has a siamese structure and jointly learns body part segmentation and label propagation on infant movement videos. 
    For more efficient semi-supervised learning, we develop two alternative learning strategies to fully utilize both labeled and unlabeled frames. 
    \item To augment the training set for body parsing, we introduce a novel factorized video GAN to synthesize new labeled video frames by composing different synthesized foregrounds (\textit{i.e.}, infant) and backgrounds.
    
\end{itemize}

In the remainder of this paper, we first conduct literature review and discuss related works in Section \ref{sec:relatedwork}; we then introduce our proposed SiamParseNet (SPN) and augmented training with factorized video GAN (FVGAN) in Section \ref{sec:Methods}. Experimental evaluation for infant body parsing and pose estimation using two datasets is presented in Section \ref{sec:exp}. Section \ref{sec:gma} introduces how to apply our proposed models to GMA and compares the performance of using different combinations of input representations on a 3rd clinical dataset with GMA annotations. We finally give conclusions in Section \ref{sec:conclusion}.

\section{Related Work}
\label{sec:relatedwork}
\subsection{
Automated General Movement Assessment
}
Based on different sensing modalities, existing automated GMA systems can be classified into direct and indirect sensing assessment {
\citep{Marcroft14, irshad2020ai}}. Direct sensing means infant movements are captured by devices directly attached to the infant subject, such as wearable movement sensors \citep{Chen16,machireddy2017video} and magnet tracking systems \citep{Philippi14}. Indirect sensing utilizes hardwares that are integrated into the assessment environment, most of which employ video-based approaches and the videos can be captured using RGB cameras {\citep{Adde09,zhang2019online,reich2021novel}}
, Kinect \citep{hesse2018learning}
or 3D motion capture systems \citep{Meinecke06}. 

Compared with other sensing methods, the RGB camera-based approaches {\citep{Adde09,orlandi2018detection,zhang2019online,chambers2020computer,doroniewicz2020writhing,reich2021novel}} have many advantages. First, the videos can be captured conveniently at home or in hospital, requiring only a cheap camera device such as the ones in mobile phones. Second, without attached sensors that affect infant motion, the video-based approach can record more spontaneous movements of infants. 
To the best of our knowledge, \cite{Adde09} designed the first video-based automatic GMA system. They utilized background subtraction and frame differencing methods to detect CP. However, their frame differencing methods could be vulnerable to the slow movement of infants, where small differences between adjacent frames may be regarded as noise and then mistakenly removed. \cite{orlandi2018detection} developed a computer-aided GMA method which can classify infant movement videos into normal and CP. They employed a skin model for infant silhouette segmentation and large displacement optical flow (LDOF) for motion tracking. Kinematic features were then extracted and fed into several classifiers for evaluation. However, their skin model required users to manually select one skin area in advance, which may be inconvenient in practice. 
{
\cite{zhang2019online} employed U-Net~\citep{ronneberger2015u} to design an online learning framework for parsing the infant body.
\cite{chambers2020computer} extracted body poses from the IMVs of at-risk infants using OpenPose \citep{cao2017realtime} to calculate infant kinematic features. Then they utilized a Na\"ive Gaussian Bayesian Surprise metric to predict infant neuromotor risk. 
Similarly, \cite{reich2021novel} also used OpenPose \citep{cao2017realtime} to estimate the infant skeleton from IMVs and then employed a shallow multi-layer neural network to classify infant motor functions. 
However, most of these frameworks simply take a video as a set of multiple frame images and do not effectively utilize the temporal relationships between continuous frames. {
Very recently, \cite{cao2022aggpose} proposed a Deep Aggregation Vision Transformer (AggPose) for infant pose estimation. Though achieving promising results on their infant pose dataset, their model is only designed for image pose estimation and does not exploit temporal continuity between video frames. The pose estimation result was also not used for CP prediction.}
}

\subsection{Zero-shot Video Object Segmentation}
Parsing infant body in IMVs is highly relevant to the video object segmentation task (VOS). According to whether requiring the object mask during testing, VOS can be classified into zero-shot solution (no annotation given for testing frames) and one-shot solution (given the label of one testing frame) \citep{ventura2019rvos}. In this paper, we only focus on reviewing zero-shot VOS methods. 
Though VOS has been widely explored for natural scenes, challenges arise when directly applying existing VOS methods to infant body parsing, due to frequent occlusion among body parts during infant movements.
Among current state-of-the-art methods, \cite{lu2019see} introduced a CO-attention Siamese Network (COSNet) to capture the global-occurrence consistency of objects of interest among video frames for VOS by taking a pair of frames from the same video as input and learn to extract their rich correlations.
However, such co-attention module may not work well for IMVs since IMVs generally use fixed camera so that the global-occurrences exist in both infant and background regions. The fixed camera characteristic of IMVs makes the global-occurrence consistency feature less distinguishable between infants and background.
\cite{ventura2019rvos} proposed a recurrent network RVOS to integrate spatial and temporal domains for VOS by employing uni-directional convLSTM~\citep{xingjian2015convolutional}. This convLSTM only considers previous frames, which may result in unsatisfactory performance when occlusion persistently appears.
\cite{zhu2017deep} proposed Deep Feature Flow, 
which first runs a convolutional sub-network on key frames for image recognition and then propagates their deep feature maps to other frames via optical flow.
Also using optical flow, in the medical video field, \cite{jin2019incorporating} proposed MF-TAPNet for instrument segmentation in minimally invasive surgery videos, which incorporates motion flow based temporal prior with an attention pyramid network. 
Optical flow based methods aim to find point-to-point correspondences between frames and have achieved promising results. However, for infant movement videos with frequent occlusions, it can be challenging to use optical flow to track corresponding points around occluded body parts. 
Moreover, few previous methods have investigated the semi-supervised training setting. As annotating IMV frames for body parsing is costly, semi-supervised methods that utilize partially labeled IMVs for training have great potential and deserve further research. 

\begin{figure*}[t]
    \centering
    \includegraphics[width=0.8\linewidth]{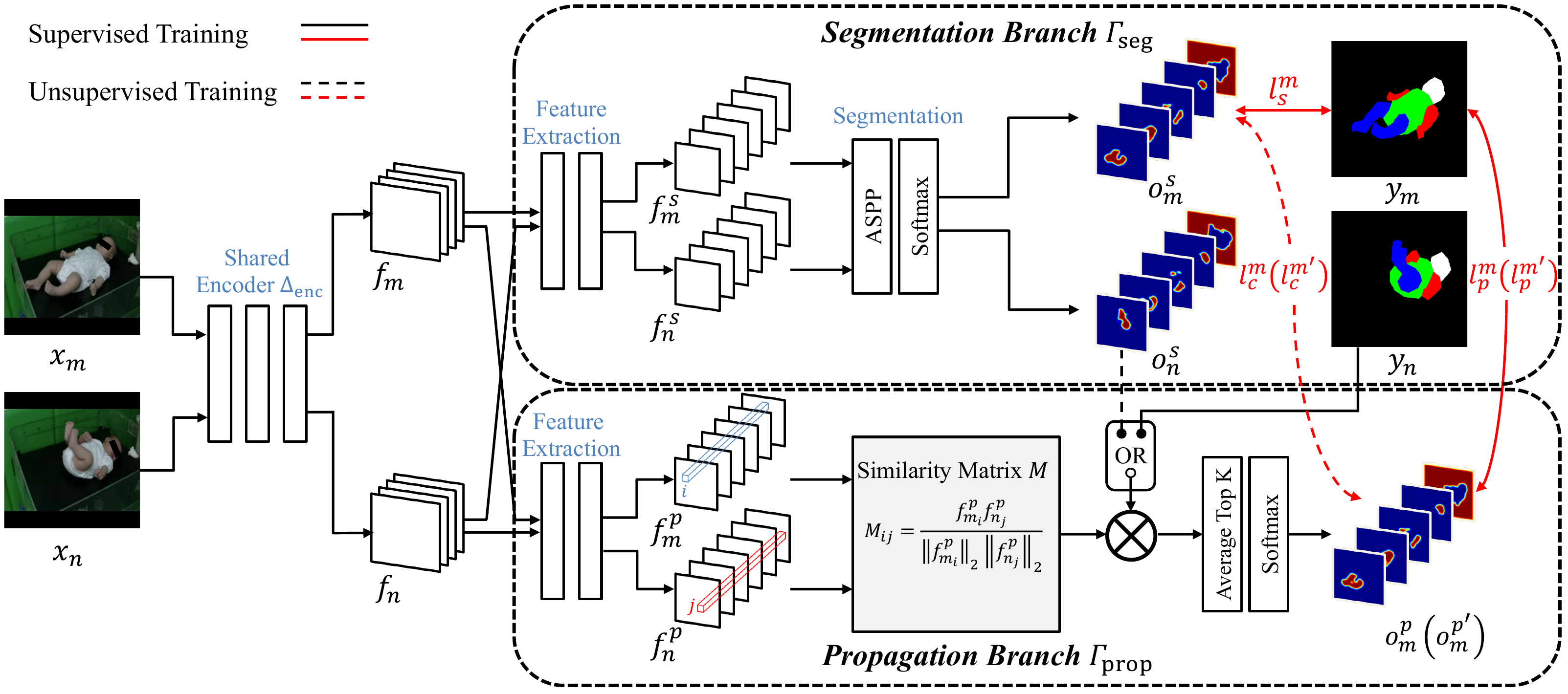}
    \caption{The training framework of SiamParseNet (SPN). $x_m$ and $x_n$ are a pair of video frames from the same video, $y_m$ and $y_n$ are corresponding annotated body masks (if available). $f$ denotes extracted visual features and $o$ refers to output probability map.}
    \label{fig:spn}
\end{figure*}

\subsection{
Guided Person Image Generation
}
To mitigate the high cost of manual annotation, we propose factorized video GAN (FVGAN) to synthesize labeled video frames based on annotated IMVs, which is related to guided image-to-image translation, especially person image generation. 
Currently, most methods of person image generation are pose-guided. 
\cite{ma2017pose} proposed a two-staged coarse-to-fine network $\rm{PG}^2$ that allows synthesizing person images in an arbitrary pose when given the image of that person and a novel pose. 
\cite{ma2018disentangled} further proposed a novel, two-stage reconstruction framework to learn a disentangled representation of person images via decomposing such images into three factors: foreground, background and pose information, and then generated novel images by manipulating these image factors.
\cite{song2019unsupervised} addressed the task of pose-guided person image generation via leveraging semantic generator and appearance generator. 
Cycle-consistency loss \citep{zhu2017unpaired} is adopted to enable training without paired images.
Despite such progress, few existing person image generation approaches explored body-parsing-mask-guided generation, which is more challenging than posed-guided generation since the shapes of body parsing masks between different subjects can be quite dissimilar while poses tend to be more subject-independent and thus pose transfer between different subjects can be easier.
Furthermore, compared with the datasets used in existing works such as DeepFashion \citep{liu2016deepfashion} and Market-1501 \citep{zheng2015scalable}, IMV datasets usually contain videos with much more complex backgrounds thus synthesizing IMV image frames is more challenging. 

\subsection{Differences from our previous work}

A preliminary version of this work is presented in our conference publication~\citep{Ni2020SiamParseNetJB}.
In~\citep{Ni2020SiamParseNetJB}, we proposed SiamParseNet (SPN), a siamese structured neural network \citep{bertinetto2016fully} for infant body parsing which takes an arbitrary pair of frames from the same video as input during training. 
It includes one shared encoder $\Delta_{\text{enc}}$ and two branches: intra-frame body part segmentation branch $\Gamma_{\text{seg}}$, and inter-frame label propagation branch $\Gamma_{\text{prop}}$, where
$\Gamma_{\text{prop}}$ is particularly designed to consider multiple possible correspondences for alleviating the frequent occlusion challenge. To jointly train the two branches $\Gamma_{\text{seg}}$ and $\Gamma_{\text{prop}}$, we also introduced a consistency loss to provide extra regularization. Since our proposed label propagation branch $\Gamma_{\text{prop}}$ can propagate either from ground truth masks or $\Gamma_{\text{seg}}$ outputs, we further utilize the consistency loss to enable semi-supervised learning (SSL) with both labeled and unlabeled video frames. To control the alternate training process between using annotated frames and unannotated ones, we proposed adaptive alternative training (AAT) in~\citep{Ni2020SiamParseNetJB}. During testing, a multi-source inference (MSI) mechanism is proposed, which combines both segmentation branch $\Gamma_{\text{seg}}$ and propagation branch $\Gamma_{\text{prop}}$ to perform body part segmentation. MSI first automatically selects \textit{key} frames and then employs branch $\Gamma_{\text{seg}}$ to semantically segment them. Subsequently, MSI utilizes branch $\Gamma_{\text{prop}}$ to propagate key-frame segmentation results to other \textit{non-key} frames. Instead of only considering the previous or current frames like most previous video object segmentation methods \citep{lu2019see,ventura2019rvos}, MSI leverages the local context provided by representative key frames in each video clip to perform parsing, which further alleviates the frequent occlusion issue in IMVs. 
{
In this work, we improve the SiamParseNet method by comparing and evaluating two alternative semi-supervised training strategies:
thresholded automated training (TAT) and adaptive alternative training (AAT). Furthermore, inspired by the recent successful application of generative adversarial networks (GANs) \citep{goodfellow2014generative} in both natural image generation \citep{vondrick2016generating,ma2018disentangled,tulyakov2018mocogan,song2019unsupervised} and medical image synthesis \citep{xue2019synthetic,xue2021selective,tajbakhsh2020embracing},  we propose a novel factorized video GAN (FVGAN) to synthesize labeled frames to augment our training dataset and reduce the burden of manual annotation. 
More specifically, FVGAN decomposes an IMV frame into two factors: foreground (infants) and background, and decouples the generation of foreground and background.  Given a body part segmentation mask from any existing labeled video frame, FVGAN can generate new video frames with various foregrounds and backgrounds (Fig.~\ref{fig:fvgan_res2}). 
 }
 
{
While only body parsing is demonstrated and evaluated using one small clinical IMV dataset~\citep{zhang2019online} in our previous work~\citep{Ni2020SiamParseNetJB}, in this work, to better assist with GMA and verify the generalizability of our proposed framework, we show that SPN can be extended to the task of infant pose estimation by solely switching its backbone network. Our proposed models for body parsing and pose estimation are validated using two datasets: (a) the BHT dataset~\citep{zhang2019online} (used in our previous work) with approximately 6\% training frames annotated and, (b) the Youtube-Infant dataset~\citep{chambers2020computer} with around 30\% labeled frames.
Body parsing results show that our proposed models can achieve comparable or better performance when compared with several state-of-the-art image/video segmentation methods~\citep{ronneberger2015u,chen2018encoder,lu2019see,ventura2019rvos, chen2020naive}. Examples of comparison between our best-performing method (SPN + FVGAN) and other methods can be found in Fig.~\ref{fig:sota}. 
We also conduct various ablation studies to prove the effectiveness of our proposed modules, including joint training of two branches, consistency loss, alternative semi-supervised training process, factorized video GAN and multi-source inference. 
Pose estimation results on an IMV dataset~\citep{chambers2020computer} demonstrate the promising performance of SPN for infant pose estimation (99.21\% PCK@0.1). 
}

{
Finally, in this work, to validate the clinical value of our proposed SPN, we propose a convolution-recurrent neural network (CRNN)-based IMV classification model for GMA and experiment on a newly collected clinical IMV dataset with expert GMA annotations. Results demonstrate that both body parsing and pose estimation results generated by SPN without any fine-tuning can significantly improve CP prediction performance on the new dataset, increasing 
the mean AUC score by more than 15\% compared to models using only raw video inputs.
}

\section{Methodology}
\label{sec:Methods}
Fig.~\ref{fig:spn} shows the overview of our proposed SiamParseNet (SPN). SPN is a siamese structure taking an arbitrary pair of training frames from the same video as input regardless of the availability of their annotation. This increases the amount of training data as well as enables the utilization of partially labeled videos. More specifically, we consider three cases of using training frame pairs according to their annotation: the fully supervised mode, which uses two labeled frames for training; the semi-supervised mode, where only one of the two input training frames is annotated; the unsupervised mode, where neither training frames is labeled. During training, we propose two alternative training strategies, thresholded automated training (TAT) and adaptive alternative training (AAT), both of which mainly rely on fully supervised mode at early stages, and rely more on semi-supervised and unsupervised modes at later stages. 
To augment the set of labeled video frames, we also propose factorized video GAN (FVGAN) to generate new labeled video frames based on existing labeled training frames. FVGAN generates a new frame by synthesizing the foreground and background separately (Fig.~\ref{fig:g}). {This helps reduce training difficulty and model complexity as well as makes it easier to apply image transformation.} 
Given a labeled video frame from one video and another randomly-chosen labeled frame from a different video, FVGAN can generate three new labeled frames to help model training (Fig.~\ref{fig:fvgan_res2}).
During testing, we propose multi-source inference (MSI) to achieve robust body parsing by combining both the segmentation branch and the propagation branch of SPN (see Fig.~\ref{fig:msi}). Next, we introduce our proposed SPN and its semi-supervised learning mechanism, alternative training strategies, factorized video GAN and multi-source inference. 

\subsection{Semi-supervised Learning}
\label{subsec:ssl}
As illustrated in Fig.~\ref{fig:spn}, given a pair of input frames $x_m$ and $x_n$, we first employ a shared encoder $\Delta_{\text{enc}}$ to extract their feature maps $f_m$ and $f_n$. Then we feed them to segmentation branch $\Gamma_{\text{seg}}$ and propagation branch $\Gamma_{\text{prop}}$. $\Gamma_{\text{seg}}$ further represents $f_m$ and $f_n$ as $f^s_m$ and $f^s_n$ and generates the segmentation probability maps $o^s_m$ and $o^s_n$ with a segmentation module. $\Gamma_{\text{prop}}$ processes $f_m$ and $f_n$ as $f^p_m$ and $f^p_n$ to calculate the similarity $M$ in the feature space. $\Gamma_{\text{prop}}$ outputs the probability maps $o^p_m$ and $o^p_n$ through different paths according to the availability of the annotation of $x_m$ and $x_n$.

\noindent\textbf{Case 1:} When both input frames $x_m$ and $x_n$ have ground truth masks available, $y_m$ and $y_n$, we have the fully-supervised mode. In this case, $\Gamma_{\text{prop}}$ propagates the ground truth mask of one frame to another, as the solid line in Fig. \ref{fig:spn} shows. 
The overall loss $l_\text{sup}$ is calculated as:
\begin{equation}
    \label{eq:l_sup}
    l_\text{sup} = l^{m}_{s} + l^{n}_{s} + l^{m}_{p} + l^{n}_{p} 
    + \lambda\left(l^{m}_{c}+l^{n}_{c}\right)\enspace,
\end{equation}
\noindent where all losses are cross-entropy losses between one-hot vectors. More specifically, $l^{m}_{s}$ and $l^{n}_{s}$ are segmentation losses of $\Gamma_\text{seg}$ between $o^s_m$ and $y_m$, $o^s_n$ and $y_n$, respectively. $l^{m}_{p}$ and $l^{n}_{p}$ are losses of $\Gamma_\text{prop}$ between $o^{p}_m$ and $y_m$, $o^{p}_n$ and $y_n$.
The consistency loss $l_{c}$ measures the degree of overlapping between outputs of the two branches $\Gamma_\text{seg}$ and $\Gamma_\text{prop}$, 
where $\lambda$ is a scaling factor to ensure $l_{c}$ have roughly the same magnitude as $l_{s}$ and $l_{p}$.

\noindent\textbf{Case 2:} If neither input frames $x_m$ and $x_n$ is labeled, we have the unsupervised mode. In this case, branch $\Gamma_{\text{prop}}$ propagates the segmentation output of one frame to another, as the dotted line in Fig. \ref{fig:spn} shows. More specifically, $\Gamma_{\text{prop}}$ transforms the output of $\Gamma_{\text{seg}}$, $o^s_m$ and $o^s_n$, to probability map $o^p_m$ and $o^p_n$. Due to the lack of ground truth, we only consider the consistency loss between two branches:
\begin{equation}
    \label{eq:l_un}
    l_\text{un}= \lambda\left(l^{m'}_{c}+l^{n'}_{c}\right)\enspace.
\end{equation}

\noindent\textbf{Case 3:} If only one in the pair of input frames is annotated, we have the semi-supervised mode. Without loss of generality, assume that $y_m$ is available but $y_n$ is not. Then branch $\Gamma_{\text{prop}}$ generates the probability map $o^{p'}_{m}$ with $\Gamma_{\text{seg}}$'s output $o^s_n$ instead of $y_n$ (the dotted line in Fig. \ref{fig:spn}).
We compute the losses $l^{m'}_{p}$ and $l^{m'}_{c}$, which measure the loss between $o^{p'}_{m}$ and $y_m$, $o^{p'}_{m}$ and $o^s_m$, respectively.
We also calculate $l^{m}_{s}$ and $l^{n}_{c}$, which measure the loss between $o^s_m$ and $y_m$, $o^p_n$ and $o^s_n$, respectively. 
Thus, the semi-supervised loss is:
\begin{equation}
    \label{eq:l_semi}
    l_\text{semi}= l^{m}_{s} + l^{m'}_{p} + \lambda\left(l^{m'}_{c}+l^{n}_{c}\right)\enspace.
\end{equation}
\vspace{-10pt}


As a general framework, SPN can adopt various networks as backbone. 
We follow DeepLab \citep{chen2017deeplab,chen2018encoder} and use the first three residual blocks of ResNet101 \citep{he2016deep} as encoder $\Delta_{\text{enc}}$. For branch $\Gamma_{\text{seg}}$, we employ the $4^{th}$ block of ResNet101 for feature extraction and ASPP \citep{chen2017deeplab} module for segmentation. Branch $\Gamma_{\text{prop}}$ also utilizes the $4^{th}$ block of ResNet101 to extract feature. Note that the weights are not shareable between the two branches and thus the two branches are optimized separately during training.

To propagate a given source segmentation map to a target frame, similar to~\cite{hu2018videomatch}, we first calculate the cosine similarity matrix $M$ of $f^{p}_{m}$ and $f^{p}_n$ as 
\begin{equation}
    \label{eq:cos}
    M_{ij}= \frac{f^{p}_{m_i} f^{p}_{n_j}}{\left\|f^{p}_{m_i}\right\|_{{2}}\left\|f^{p}_{n_j}\right\|_{{2}}}\enspace,
\end{equation}
where $M_{ij}$ is the affinity value between $f^{p}_{m_i}$, point $i$ in map $f^{p}_{m}$, and $f^{p}_{n_j}$, point $j$ in map $f^{p}_n$.
Then, given the source segmentation map $\hat{y}_n$ (either ground truth $y_n$ or $\Gamma_\text{seg}$ generated $o^s_n$), and the similarity matrix $M$, $\Gamma_\text{prop}$ produces $o^{p}_{m_i}$, point $i$ in output map $o^{p}_m$ as
\begin{equation}
    \label{eq:sim}
    o^{p}_{m_i} = \text{softmax}\left(\frac{1}{K}\sum\nolimits_{j\in \text{Top}(M_i, K)} M_{ij}\hat{y}_{n_j}\right)\enspace,
\end{equation}
where $\text{Top}(M_i, K)$ contains the indices of the top $K$ most similar scores in the $i^{th}$ row of $M$. 
Since $\Gamma_\text{prop}$ considers multiple correspondences for a point rather than one-to-one point correspondences as in optical flow \citep{zhu2017deep,meister2018unflow,jin2019incorporating}, SPN can naturally better handle occlusions in IMVs than optical flow based methods. 

\subsection{Alternative Training}
As mentioned briefly in the beginning of Section \ref{sec:Methods}, during the training of SPN, we adopt two training strategies, thresholded automated training (TAT) and adaptive alternative training (AAT), to alternatively employ different training modes to achieve optimal performance.
{Intuitively, SPN should rely more on the supervised mode at early stages and then gradually incorporate more semi-supervised and unsupervised training at later stages of training.} 
To dynamically adjust the reliance on different training modes, we propose TAT and AAT to automatically sample training data among the three cases. Assume that the probabilities of selecting case 1, case 2, and case 3 for any iteration/step of training are $p_1$, $p_2$, and $p_3$, respectively. Considering case 2 and case 3 both involve utilizing unlabeled frames, we set $p_2 = p_3 = \frac{1-p_1}{2}$. 
Thus we only need to control the probability of choosing case 1 training and the other two cases are automatically determined. 

\begin{figure*}[t]
    \centering
    \includegraphics[width=0.80\linewidth]{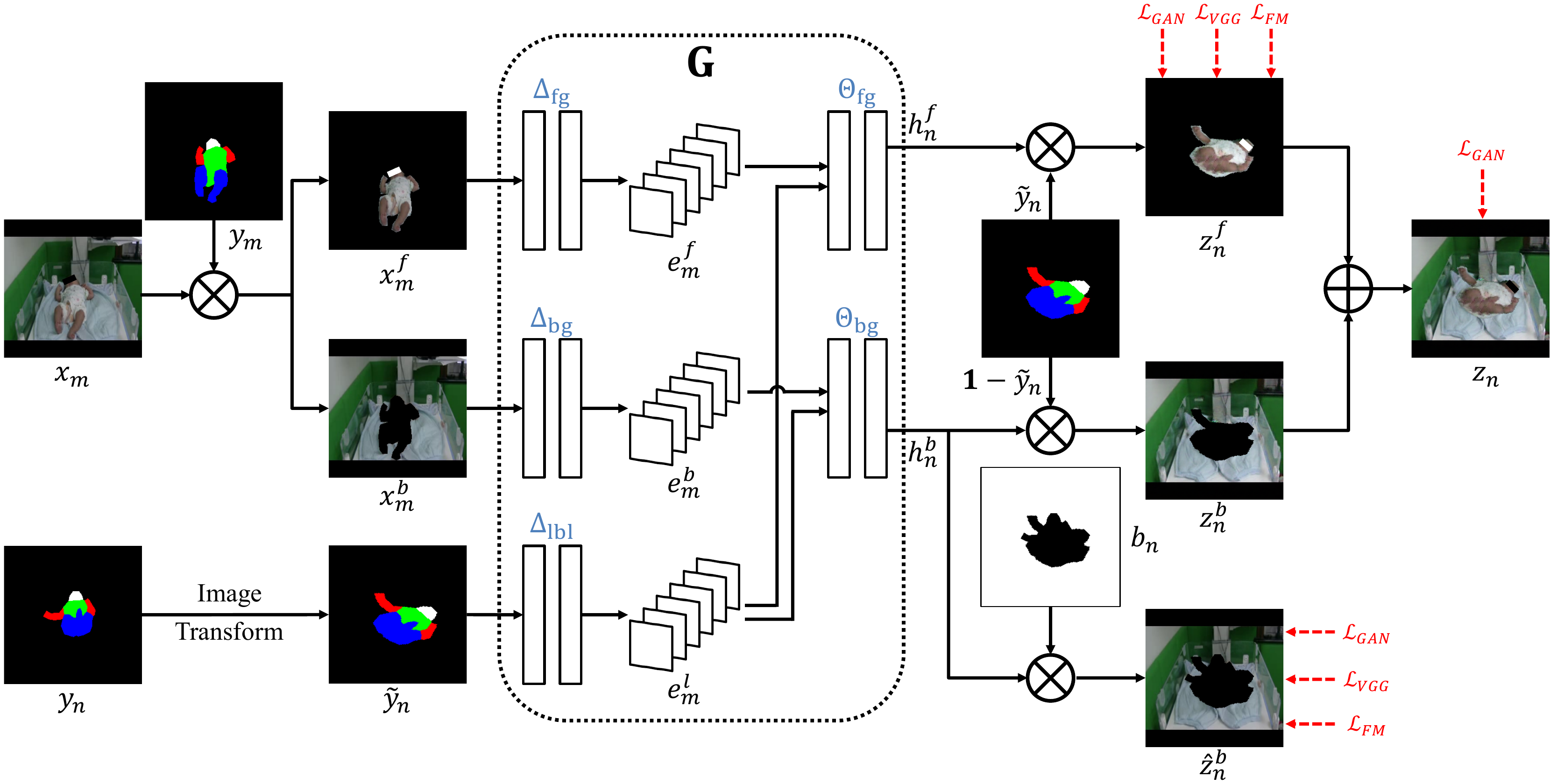}
    \caption{The training framework of our proposed factorized generator $G$. In general, $x$ refers to real video frame, $y$ refers to original body parsing mask, $\tilde{y}$ is the transformed mask, and $z$ is synthesized video frame. Superscripts $f, b$ correspond to foreground infant body and background image. Subscripts $m, n$ indicate two input sources from the same video. Note that $b_n=(\mathbf{1}-y_n)\odot (\mathbf{1}-\tilde{y}_n)$.}
    \label{fig:g}
\end{figure*}

For TAT, we change $p_1$ according to the pixel accuracy $\epsilon$ during training, where $p_1$ is calculated as:
\begin{equation}
    \label{eq:thre}
    p_1 = \left\{
    \begin{aligned}
    1,\quad &\text{if}\  \epsilon<\mathbb{T}\\
    \mathbb{P}_t,\quad &\text{if}\ \epsilon\geq \mathbb{T}
    \end{aligned}
    \right.
\end{equation}
Since the moving average pixel accuracy of the output of branch $\Gamma_{\text{seg}}$ may differ from that of branch $\Gamma_{\text{prop}}$, we choose $\epsilon$ as the minimum of the two accuracy values.
$\mathbb{T}$ is the preset accuracy threshold and $\mathbb{P}_t$ is the preset probability. Note that $\epsilon$ is continuously changing during training. When $\epsilon$ is lower than the preset $\mathbb{T}$, we only adopt the fully-supervised mode as $p_1 = 1$.

For AAT, we use an annealing temperature $t$ to gradually reduce $p_1$ as training continues, where $p_1$ is computed as
\begin{equation}
    \label{eq:ada}
        p_1 = \text{max}\left(1-
        \left(
        1-\mathbb{P}_1
        \right)
        \left(\frac{i}{i_\text{max}}\right)^t, \mathbb{P}_1\right)\enspace,
\end{equation}
where $i$ is the training step, $\mathbb{P}_1$ is the pre-defined lower bound probability of $p_1$, $i_\text{max}$ is the maximum number of steps of using AAT.

\subsection{
Augmented Training with Factorized Video GAN
}
To alleviate the high cost of manual annotation and use synthetic data to augment the set of labeled training frames, we propose factorized video GAN (FVGAN) to generate synthetic labeled IMV frames. Given an input training IMV frame and its corresponding mask $(x_m,y_m)$, and the target body parsing mask $y_n$, the generator $G$ of FVGAN can output a new frame $z_n$ which is a frame of the same infant and background scene as $x_m$ but with the body parsing mask as $y_n$, that is, 
\begin{equation}
\label{eq:fvgan}
    G((x_m, y_m), y_n) = z_n\enspace,
\end{equation}
this synthesized labeled frame and its mask ($z_n$, $y_n$) can be later added to the original training set for data augmentation.

When training FVGAN, we randomly sample two labeled frames ($x_m$, $y_m$) and ($x_n$, $y_n$) from the same video for supervised learning. Here $x_m$ and $x_n$ are a pair of images which have the same foreground (infant appearance) and background scene but are under different body parsing masks $y_m$ and $y_n$. Using labeled frames from the same video makes it possible to use both adversarial and supervised losses for training.
 When synthesizing new frames using FVGAN, we randomly select two masks $y_m$ and $y_n$ from different videos, and then we utilize frame $x_m$ to generate new frame $z_n$ as in {Eq.~\ref{eq:fvgan}}. By choosing $y_n$ from a video that is different from that of $y_m$, it is guaranteed that the generated frame is a novel frame.

To simulate {cross-video} inference as well as reduce model overfitting, when training FVGAN, we randomly apply several large-degree image transformation operations to the original target mask $y_n$, including rotation, translation, scaling, and shear. 
To leverage supervised training, we also apply the same image transformation to the target frame $x_n$. 
However, different from the traditional image transformations which act on the whole image, transformations are only applied to the foreground (infant body) part of $x_n$, $x^f_n$, to avoid unnecessary changes to the background scene.
That is, we apply the same transformation as $y_n$ to $x^f_n$. For background regions, applied transformations only affect those areas occluded by the original or transformed foreground, where all remaining regions in the background are kept unchanged. 

Due to such transformation difference between foreground and background, 
we decouple the generation of foreground and background in FVGAN, where foreground synthesis and background synthesis have independent training losses. The factorized design of FVGAN eases the manipulation and enables diverse synthesis outputs by combining the foreground and background of the input frame with synthesized foreground and background. Overall, FVGAN can generate three types of images, including synthesized foreground plus target background image (termed $I_{\text{fore}}$), synthesized background plus target foreground image (termed $I_{\text{back}}$), synthesized foreground and synthesized background image (termed $I_{\text{whole}}$).


The detailed structure of the generator $G$ in FVGAN is illustrated in Fig.~\ref{fig:g}.
In general, we first factorize input frame $x_m$ to foreground embedding feature $e^f_m$ and background feature $e^b_m$, and then combine them with $e^l_n$ which is the label feature of target body parsing mask, to generate the synthesized foreground image $z^f_n$ and background image $z^b_n$, respectively. We finally generate $z_n$ by assembling $z^f_n$ and $z^b_n$ together.

More specifically, during training, for the input image pair $(x_m, y_m)$, 
we first utilize the mask $y_m$ to decompose the image $x_m$ into foreground infant body image $x^f_m$ and background scene image $x^b_m$.
To promote diversity in body masks,
we apply large-degree random transformation to target mask $y_n$ and get transformed mask $\tilde{y}_n$. 
Then, we adopt three feature encoders $\Delta_{\text{fg}}$, $\Delta_{\text{bg}}$, and $\Delta_{\text{lbl}}$ to encode foreground image $x^f_m$, background image $x^b_m$, and the mask $\tilde{y}_n$ as feature $e^f_m$, $e^b_m$ and $e^l_n$, respectively. We subsequently feed foreground feature $e^f_m$ and target label feature $e^l_n$ to the foreground decoder $\Theta_{\text{fg}}$, and feed the background feature $e^b_m$ and label feature $e^l_n$ to the background decoder $\Theta_{\text{bg}}$. We further use their output $h^f_n$ and $h^b_n$ to generate the synthesized foreground image $z^f_n$ and background image $z^b_n$, where $z^f_n=h^f_n \odot \tilde{y}_n$, $z^b_n=h^b_n \odot (\mathbf{1}-\tilde{y}_n)$. Here $\odot$ is pixel-wise multiplication and $\mathbf{1}$ is a matrix of ones. We finally sum up $z^f_n$ and $z^b_n$ to obtain the final synthetic image $z_n$. 

Note that the ground truth of $z_n$ is missing after employing image transformation.
To enable supervised training, we additionally generate new ground truth masks for synthesized foreground $z^f_n$ and background $z^b_n$. We apply the same image transformation as $y_n$ to the foreground of the target image $x^f_n$ and get $\tilde{x}^f_n$ as the supervision signal of $z^f_n$. For $z^b_n$, it is hard to estimate background areas covered by $y_n$ but not by $\tilde{y}_n$. We choose to ignore such ambiguous areas by creating a background mask $b_n$, where $b_n=(\mathbf{1}-y_n)\odot (\mathbf{1}-\tilde{y}_n)$. Then we generate the combined background image $\hat{z}^b_n$, where $\hat{z}^b_n=h^b_n\odot b_n$, and its corresponding supervision signal $\hat{x}^b_n$, where $\hat{x}^b_n=x_n\odot b_n$.

Our proposed encoder $\Delta$ and decoder $\Theta$ in FVGAN can use various backbone networks, such as the ones in DCGAN~\citep{radford2015unsupervised} and pix2pixHD~\citep{wang2018high}. Here we adopt the architecture similar to pix2pixHD due to its strong representation power. For the encoder $\Delta_{\text{fg}}$ and $\Delta_{\text{bg}}$, we use the network with four stride-2 convolutions and 9 residual blocks. For the encoder $\Delta_{\text{lbl}}$, we only use four stride-2 convolutions without additional residual blocks because mask images are simple. For both decoder $\Theta_{\text{fg}}$ and $\Theta_{\text{bg}}$, we employ four fractionally-strided convolutions with stride $\frac{1}{2}$. Similar to \cite{johnson2016perceptual}, we use instance normalization~\citep{ulyanov2016instance}. For the discriminator $D$, we use $70\times 70$ PatchGANs~\citep{zhu2017unpaired,isola2017image,wang2018high}, which aims to classify whether the $70\times 70$ overlapping patches are real or fake. To stabilize the training, we use LSGAN~\citep{mao2017least} as the adversarial loss.

The overall training loss of FVGAN $l_{\text{FV}}$ is defined as:
\begin{equation}
   l_{\text{FV}} = l_{\text{whole}} +  l_{\text{fore}} + l_{\text{back}}
   \enspace,
\end{equation}
where the whole image loss $l_{\text{whole}}=\mathcal{L}_{\text{GAN}}(x_n, z_n)$, the foreground image loss $l_{\text{fore}}=\mathcal{L}(\tilde{x}^f_n, z^f_n)$, and the background image loss $l_{\text{back}}=\mathcal{L}(\hat{x}^b_n, \hat{z}^b_n)$. 
The definition of $\mathcal{L}(x, z)$ is:
\begin{equation}
    \label{eq:lxo}
    \mathcal{L}(x, z) = \mathcal{L}_{\text{GAN}}(x, z) + \gamma\mathcal{L}_{\text{VGG}}(x, z) + \beta \mathcal{L}_{\text{FM}}(x, z)\enspace,
\end{equation}
where $\gamma$ and $\beta$ are scaling factors, and the adversarial loss $\mathcal{L}_{\text{GAN}}(X,Y)$ is defined by the following minimax game \citep{goodfellow2014generative}:
\begin{equation}
    \min_{G}\max_{D}~\mathbb{E}_{X}[\log D(X)] + \mathbb{E}_{Y}[\log(1-D(Y))]\enspace,
\end{equation}
where $Y$ is associated with the generator $G$. We design different discriminators for different components.
For $l_{\text{whole}}$, the discriminator $D$ is designed to distinguish the real image $x_n$ from the generated image $z_n$. 
For $l_{\text{fore}}$, $D$ aims to distinguish the real image pair $(\tilde{x}^f_n, \tilde{y}_n)$ and the generated image pair $(z^f_n, \tilde{y}_n)$. For $l_{\text{back}}$, $D$ tries to differentiate $(\hat{x}^b_n, b_n)$ from $(\hat{z}^b_n, b_n)$.
The VGG loss $\mathcal{L}_{\text{VGG}}(X, Y)$~\citep{wang2018high} is defined as
\begin{equation}
    \sum^{N}_{i=1}\frac{1}{M_i}\left[||F^{(i)}(X)-F^{(i)}(Y)||_{1}\right]\enspace,
\end{equation}
where $N$ is the number of layers in VGG feature extraction network and $F^{(i)}$ denotes the output of $i$-th layer with $M_{i}$ elements of the VGG network~\citep{simonyan2014very} pretrained on ImageNet~\citep{deng2009imagenet}. The feature matching loss $\mathcal{L}_{\text{FM}}(X, Y)$ \citep{wang2018high} is defined as
\begin{equation}
    \sum^{T}_{i=1}\frac{1}{N_i}\left[||D^{(i)}(X)-D^{(i)}(Y)||_{1}\right]\enspace,
\end{equation}
where $D^{(i)}$ denotes the $i$-th layer with $N_i$ elements of our proposed discriminator $D$.

\subsection{Multi-source Inference for Testing}
\label{sec:infer}
\begin{figure}[t]
    \centering
    \includegraphics[width=\linewidth]{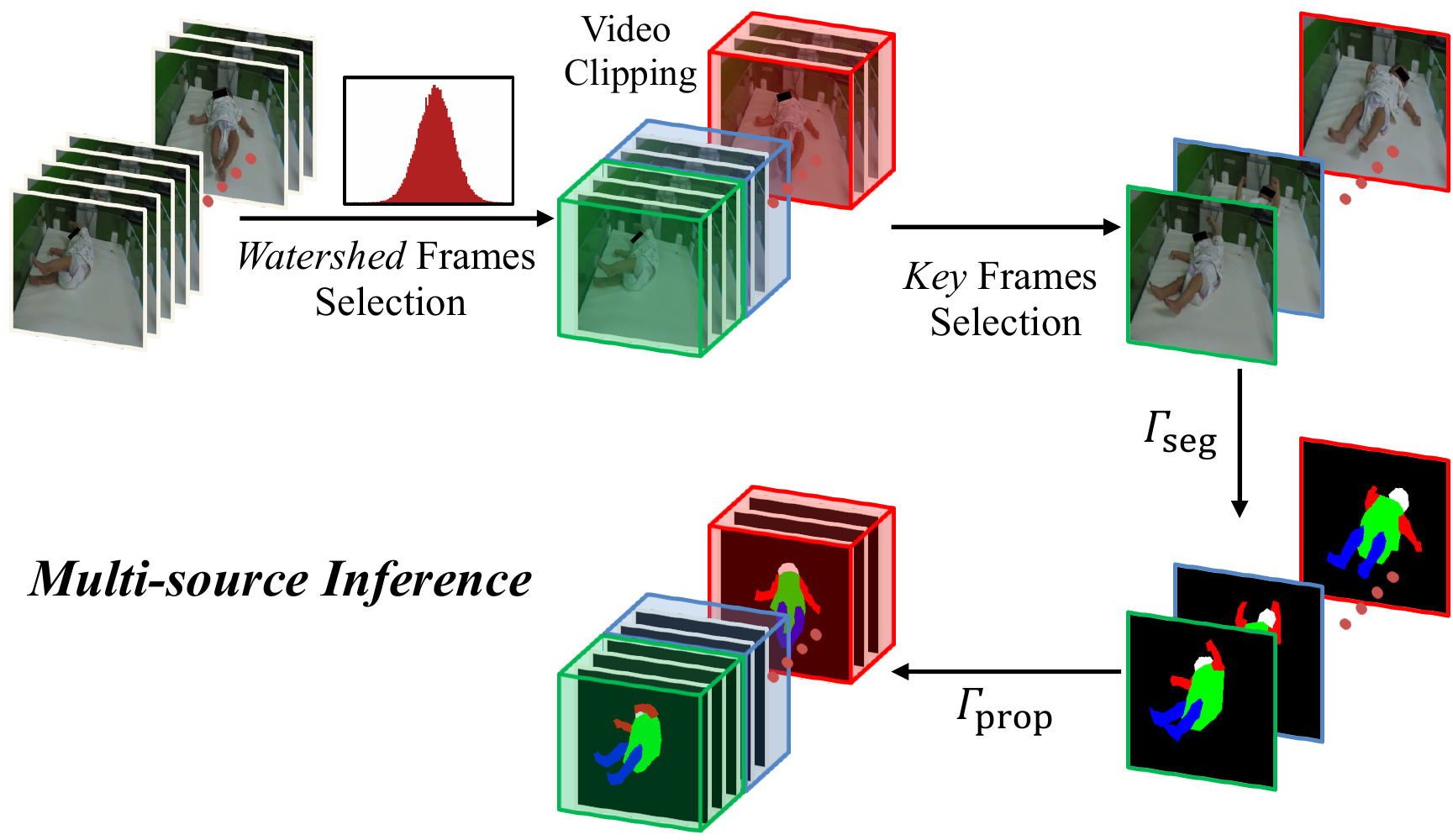}
    \caption{Multi-source inference (MSI) for testing.}
    \label{fig:msi}
\end{figure}

To further mitigate the occlusion issues in IMVs when testing, as Fig. \ref{fig:msi} shows, we propose multi-source inference (MSI) to fully exploit the dual branches of SPN. For each testing IMV, we first calculate the pixel difference between consecutive frames and then model the difference with a Gaussian distribution. The $\alpha$-th percentile is selected as threshold to sample those \textit{watershed} frames, whose pixel differences from adjacent frames are higher than the threshold. We subsequently split the IMV into several shorter clips delimited by these \textit{watershed} frames, so the infant pose and appearance is similar within each video clip. We further choose the middle frame of each clip as the $key$ frame to represent that clip because middle frame has the least cumulative temporal distance from other frames \citep{griffin2019bubblenets}. During inference, $\Gamma_{\text{seg}}$ segments the selected key frames of each video clip. Then, for the other non-key frames within the same clip, $\Gamma_{\text{prop}}$ takes the segmentation output of the corresponding $key$ frame and propagates it to other non-key frames within the clip. By splitting the long video to short clips, and using $key$ frame to provide local context and the propagation source of each clip, the proposed MSI can effectively alleviate the occlusion issues in IMVs. 
\section{Experiments for Infant Body Parsing and Pose Estimation in IMVs}
\label{sec:exp}
\subsection{
{
Datasets and Metrics
}
}
{We conduct extensive experiments on two IMV datasets: BHT dataset and Youtube-infant dataset.}

\noindent\textbf{BHT dataset}. This dataset \citep{zhang2019online} is collected through a GMA platform developed by Shenzhen Beishen Healthcare Technology Co., Ltd (BHT). 20 movement videos of infants aging from 0 - 6 months are recorded by either medical staff in hospital or parents at home. Due to the long length of original videos, we downsample them every 2 to 5 frames and the final average length of these downsampled videos is 1,500 frames. All the frames are resized to $256\times 256$ and some of them are pixel-wise annotated with five categories: background, head, arm, torso and leg. This challenging dataset covers a large variety of video contents, including diverse infant pose, varied body appearance, multiple background scenes, and viewpoint changes. We randomly split this dataset into 15 training videos and 5 testing videos, which generates that 1,267 labeled frames and 21,154 unlabeled frames in the training set (\textit{i.e.} only 5.7\% frames are labeled), and 333 labeled frames and 7,246 unlabeled frames in the testing set.

\noindent\textbf{Youtube-Infant dataset}.
We further evaluate our proposed methods on various infant videos collected from Youtube~\citep{chambers2020computer}. URLs of videos were provided by authors of~\cite{chambers2020computer}. 
For infant body parsing, we select 90 IMVs of various types from the Youtube dataset, where the average length of each video is about 120 frames. Similar to the BHT dataset, we collect annotations for five classes: background, head, arm, torso and leg.  Around 30\% of the frames for each video are manually annotated through the Amazon Mechanical Turk crowd-sourcing platform\footnote{{Data and annotations are publicly available at: \url{https://github.com/nihaomiao/Youtube-Infant-Body-Parsing}.}}. 
This Youtube infant dataset is more diverse and challenging than the BHT dataset, as it includes infants of different races, 
poses and appearances. The videos have varying background scenes, viewpoints and lighting conditions. We randomly split all videos into 68 training videos and 22 testing videos, resulting in 2,149 labeled and 4,690 unlabeled training frames, and 1,256 labeled and 2,737 unlabeled testing frames.

To better extract kinematic features and assist with GMA, we also apply SPN to pose estimation for IMVs. For pose estimation on the Youtube dataset, we follow~\cite{chambers2020computer} and choose 94 infant videos which include 84 training videos and 10 testing videos. All videos are fully labeled with 17 keypoints (nose, left/right eye, left/right ear, left/right shoulder, left/right elbow, left/right wrist, left/right hip, left/right knee, left/right ankle) by either human annotators or OpenPose \citep{cao2017realtime}. In total, there are 5,681 training frames and 487 testing frames from the videos. 

\noindent\textbf{Metrics}.
{For body parsing tasks,
we first compute the mean Dice score of all labeled testing frames for each video and then report the average score across all videos as the final result. 
When calculating Dice, we represent each pixel label of the segmentation map as a one-hot vector and ignore background pixels to focus on the body parts. We also report individual mean Dice scores for each body part.
}

{
For pose estimation tasks,
we first compute the mean PCK score \citep{yang2012articulated} of all labeled testing frames for each video and then report the average score over all the videos as the final result. The PCK score is defined as the percentage of correct keypoints, where a candidate keypoint is considered as correct if it falls within $\delta\cdot\text{max}(h, w)$ pixels of the ground truth keypoint. Here $h$ and $w$ are the height and width of the bounding box of infant body, and $\delta$ controls the relative threshold of precision.  By using two thresholds $\delta=0.1$ and $\delta=0.05$, we report two PCK scores: PCK@0.1 and PCK@0.05.
}

\subsection{Implementation details}
When training SPN for infant body parsing,
to accelerate training and reduce overfitting, similar to \cite{chen2017deeplab,chen2018encoder,lu2019see,ventura2019rvos}, we utilize the weights of DeepLab V3+ \citep{chen2018encoder} pretrained on COCO dataset \citep{lin2014microsoft} to initialize the shared encoder $\Delta_{\text{enc}}$, branch $\Gamma_{\text{seg}}$ and $\Gamma_{\text{prop}}$. For branch $\Gamma_{\text{prop}}$, we set $K=20$ in Eq.~\ref{eq:sim}, after grid searching $K$ from 5 to 20 with an interval 5. The scaling factor $\lambda$ in Eq.~\ref{eq:l_sup},~\ref{eq:l_un},~\ref{eq:l_semi} is set to be $10^{-6}$ to keep all the loss terms have the same magnitude. To construct the training frame pair from the same video for SPN, we collect different image pairs for three training modes mentioned in Section \ref{sec:Methods}. For training case 1 (\textit{i.e.} fully supervised mode), we randomly select a pair of frames of the same video from labeled training frames and repeat until we have 10,000 training image pairs, with 20,000 images in total. For training case 2 (\textit{i.e.} unsupervised mode), we repeat randomly selecting a pair of unlabeled video frames until we have 10,000 image pairs. For training case 3 (\textit{i.e.} semi-supervised mode), we repeat randomly choosing one unlabeled frame and one annotated frame until we have 10,000 image pairs. For semi-supervised learning with TAT, we set $\mathbb{P}_t=\frac{1}{3}$ in Eq.~\ref{eq:thre}. Unless otherwise specified, we set $\mathbb{T}$ to be 0.85 in all experiments. For semi-supervised learning with AAT, we set $\mathbb{P}_1 = \frac{1}{3}$ and $i_{\text{max}} = 20$ epochs. Unless otherwise specified, $t$ in Eq.~\ref{eq:ada} is set to be 0.4 in all experiments. SGD optimizer is employed with the momentum of 0.9. We set the initial learning rate to be $2.5 \times 10^{-4}$ and adopt the poly learning rate policy \citep{chen2017deeplab} with the power of 0.9. The training batch size is set to be 20. Some traditional data augmentation operations are also applied, such as color jitter, rotation and flipping. We terminate training when the pixel accuracies of both branches are almost unchanged for 2 epochs. 

To train FVGAN, similar to the fully supervised training mode of SPN, we randomly select 10,000 pairs of labeled frames from the same video in the training set. We set the batch size as 20 and train the model for 100 epochs. The Adam optimizer \citep{kingma2014adam} is employed with $\beta_1=0.5$ and $\beta_2=0.999$. The learning rate is fixed to be $2\times 10^{-4}$. We set $\gamma$ and $\beta$ in Eq.~\ref{eq:lxo} to be 10. Considering that there can be large pose difference between input mask $y_m$ and target mask $y_n$ when generating new frames since $y_n$ comes from a different video from $y_m$ during synthesis, we apply large-degree image transformations to $y_n$ during training. More specifically, the range of rotation degree, shear degree, translation pixel and scaling coefficient is $-90^{\circ}\sim90^{\circ}$, $-30^{\circ}\sim30^{\circ}$, $-25\sim25$, and $0.7\sim1.3$, respectively. 

When testing to parse infant bodies in IMVs using MSI, we set $\alpha$ in the key frame selection algorithm to be $0.98$. DenseCRF \citep{krahenbuhl2011efficient} is adopted as final post-processing. 

{
To adapt SPN to pose estimation, we only need to switch the backbone network used for body parsing to a pose estimation backbone, and replace the segmentation branch $\Gamma_\text{seg}$ with the pose estimation branch $\Gamma_\text{pose}$. 
We follow the architecture of Pose-ResNet \citep{xiao2018simple} to design the new backbone:
we first employ the four residual blocks of ResNet101 \citep{he2016deep} as shared encoder $\Delta_\text{enc}$. For the newly designed pose estimation branch $\Gamma_\text{pose}$, we employ three deconvolutional layers for feature extraction and one final convolutional layer for pose estimation. Branch $\Gamma_\text{prop}$ also utilizes three deconvolutional layers to extract features. 
Similar to the training for body parsing, we randomly select 10,000 image pairs to train a fully-supervised SPN and also utilize the weight of Pose-ResNet pretrained on COCO dataset \citep{lin2014microsoft} to initialize the model. For branch $\Gamma_\text{prop}$, we simply set $K=1$ in Eq.~\ref{eq:sim} for efficiency. 
Following the training setting in Pose-ResNet, we set the initial learning rate to be $1\times10^{-3}$ and drop it to $1\times10^{-4}$ at 5 epochs and $1\times10^{-5}$ at 10 epochs. There are 20 epochs in total. The training batch size is 20 and Adam \citep{kingma2014adam} optimizer is used. During testing, instead of using MSI, we only employ the trained pose estimation branch. We choose to only use this branch since propagating pose keypoints is rather sensitive to the selection of source frames, and MSI may fail when source frames lack required keypoints due to occlusion or mis-estimation, which is common in pose estimation tasks.
}

\subsection{Result Analysis}
\label{subsec:exp_res}

To validate the effect of joint training of $\Gamma_\text{seg}$ and $\Gamma_\text{prop}$,
we first compare two variants of SPN under fully supervised training mode: [Single-$\Gamma_\text{seg}$] which is trained using only segmentation branch and [SPN-$\Gamma_\text{seg}$] which is trained jointly but only $\Gamma_\text{seg}$ is used for all testing frames. Since $\Gamma_\text{prop}$ is not available, multi-source inference is not used for testing. 
As Table \ref{tab:semi} shows, compared with [Single-$\Gamma_\text{seg}$], joint training of SPN greatly boosts the mean dice of $\Gamma_\text{seg}$ by over $9\%$, which can be contributed to the siamese structure, the shared encoder, the consistency loss, among others.
To further validate the effect of the consistency loss, we also compare two variants of the SPN model: [SPN w/o $l_c$] which is SPN trained without using consistency loss , semi-supervised learning, and [SPN w/o SSL] which is SPN trained with consistency loss but without SSL. For both variants, MSI is used for testing.
As shown in Table \ref{tab:semi}, [SPN w/o SSL] gives better mean Dice than [SPN w/o $l_c$], which demonstrates the usefulness of the consistency loss, even in the fully supervised setting and normal training.
In addition, by comparing the results of [SPN-$\Gamma_\text{seg}$] and [SPN w/o SSL], one can see the effectiveness of MSI since the only difference between those two models is the application of multi-source inference.

\begin{table}[t]
\centering
\caption{
{
Ablation study of SPN for body parsing on BHT dataset. {Unless specially marked as [SPN w/o SSL], SPN is by default trained under the semi-supervised setting.}}
}
\label{tab:semi}
\resizebox{\linewidth}{!}{
\begin{tabular}{l|cccc|c}
\hline
Methods                  & Head           & Arm            & Torso          & Leg            & Dice           \\
\hline
Single-$\Gamma_\text{seg}$  & 67.49 & 49.49 & 73.20 & 71.78 & 69.68 \\
SPN-$\Gamma_\text{seg}$ & {79.18} & {61.00} &81.34 & 82.20 &79.49 \\
\hline
SPN w/o $l_c$ & 71.36          & 62.47          & 80.60          & 81.34          & 79.48          \\
SPN w/o SSL                 & {75.34} & {62.26} & 81.35          & 82.26          & 80.09          \\
\hline
SPN-$\Gamma_{
\text{seg}}$ ($\mathbb{T}=0.85$) & 80.03          & 62.98          & 83.38          & 82.01          & 80.54          \\
SPN-$\Gamma_{
\text{seg}}$ ($\mathbb{T}=0.90$) & 79.13          & 62.75          & 82.96          & 81.91         & 80.31          \\
SPN-$\Gamma_{
\text{seg}}$ ($\mathbb{T}=0.95$) & 78.43         & 63.49 & 81.52         & 82.93          & 80.26\\
\hline
SPN ($\mathbb{T}=0.85$) & 74.17          & 63.65          & 83.90         & 82.51          & 81.42          \\
SPN ($\mathbb{T}=0.90$) & 73.91          & 63.72          & 82.30          & 82.31          & 80.95         \\
SPN ($\mathbb{T}=0.95$) & 73.79          & {64.69} & 81.76           & 83.03         & 81.12\\
\hline
SPN-$\Gamma_{
\text{seg}}$ ($t=1.4$)   & 78.89          & 63.92          & 81.74          & 83.05          & 80.51          \\
SPN-$\Gamma_{
\text{seg}}$ ($t=0.9$)    & 78.28          & 62.93          & 81.29          & 82.34          & 79.84          \\
SPN-$\Gamma_{
\text{seg}}$ ($t=0.4$)   & 80.35          & 64.19          & 83.61 & 83.02 & 81.29 \\
\hline
SPN ($t=1.4$)    & 74.27          & 63.52          & 81.61          & 83.31         & 81.06          \\
SPN ($t=0.9$)    & 73.80          & 63.98          & 81.40          & 82.76          & 80.79         \\
SPN ($t=0.4$)   & 74.38          & 64.33          & {83.18} & {83.51} & {81.67} \\
\hline
SPN w/o SSL + FVGAN & \textbf{82.36}          & {71.22}         & \textbf{86.61}          & 84.43 & 84.13 \\
SPN ($t=0.4$) + FVGAN & 82.25         & \textbf{71.77}          & 85.12         & \textbf{85.06} &  \textbf{84.54}\\
\hline
\end{tabular}
}
\end{table}


\begin{table}[t]
\centering
\caption{
{
SPN without and with SSL trained using different numbers of completely unlabeled videos, out of the 15 training videos on BHT dataset.
}
}
\label{tab:rmv}
\resizebox{\linewidth}{!}{
\begin{tabular}{l|c|c|c|c|c|c}
\hline
\multirow{2}{*}{Methods} & \multicolumn{2}{c|}{0/15} & \multicolumn{2}{c|}{8/15} & \multicolumn{2}{c}{10/15} \\ \cline{2-7} 
                         & Dice         & Gain        & Dice         & Gain       & Dice         & Gain       \\ \hline
SPN w/o SSL                  & 80.09        & 0           & 77.17        & 0          & 71.52        & 0          \\ \hline
SPN ($\mathbb{T}=0.85$)       & 81.42        & 1.33        & 79.35        & 2.18       & 77.83       & {6.31}       \\
SPN ($t=0.4$)       & 81.67        & 1.58        & 79.21        & 2.04      & 77.83        & {6.31}       \\
 \hline
\end{tabular}
}
\end{table}

For SPN model with SSL, we experiment with different values of the pixel accuracy threshold $\mathbb{T}$ in TAT (Eq. \ref{eq:thre}) and different setting of 
the annealing temperature $t$ in AAT (Eq. \ref{eq:ada}). 
From Table \ref{tab:semi}, one can observe that semi-supervised learning using either TAT or AAT can improve the performance under different parameter settings.
To verify the effect of proposed MSI, we also try to only use the segmentation branch for testing after SSL learning. Comparing the [SPN-$\Gamma_{\text{seg}}$] with SPN using different $\mathbb{T}$ and $t$, one can find that MSI improves the final Dice for all these settings.
\begin{figure*}[t]
    \centering
    \includegraphics[width=0.75\linewidth]{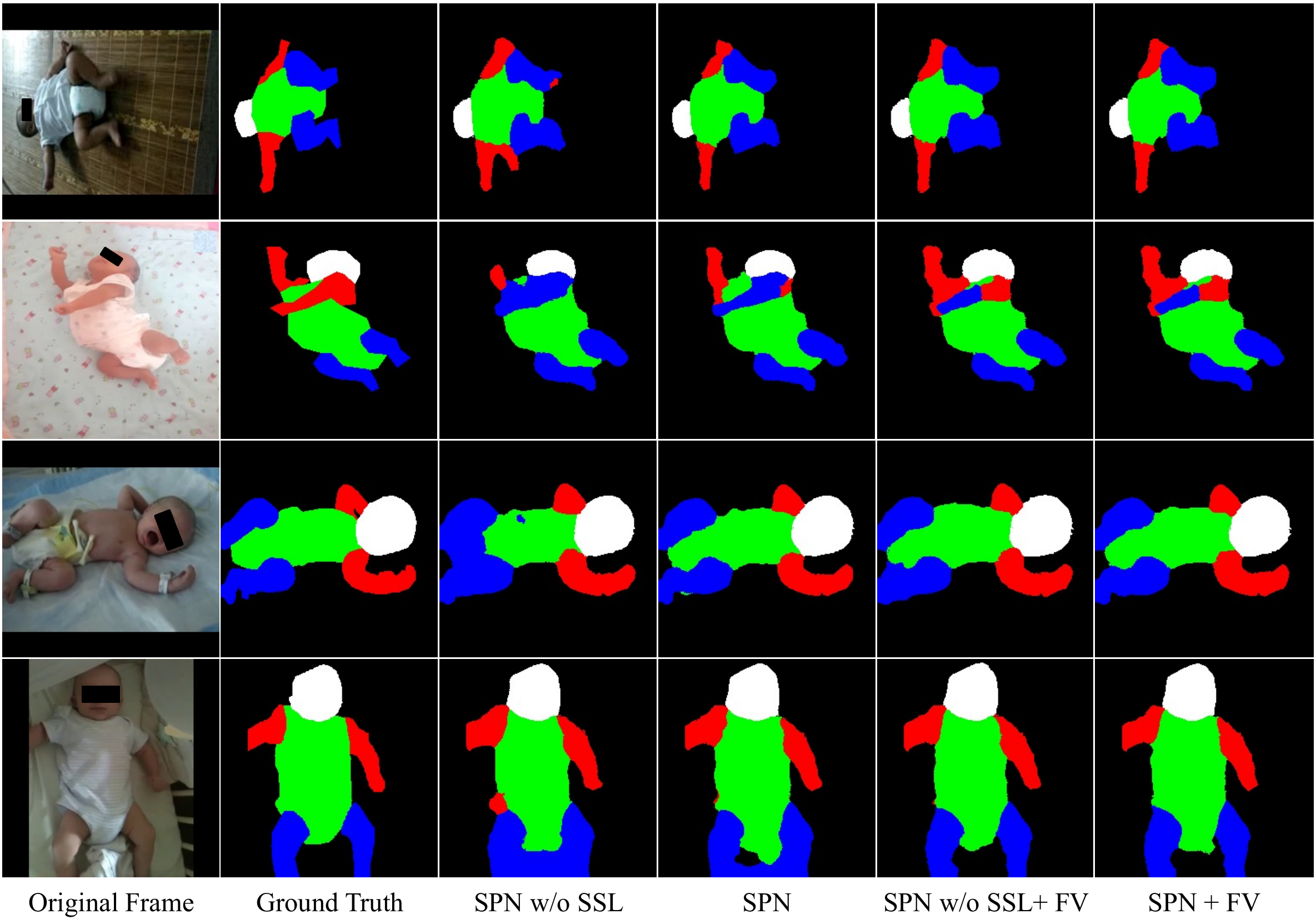}
    \caption{
    {
    Qualitative ablation study for body parsing on BHT dataset (1st and 2nd rows) and Youtube-Infant dataset (3rd and 4th rows). FV refers to models trained with FVGAN generated images.}
    }
    \label{fig:ablation}  
\end{figure*}

\begin{figure*}[t]
    \centering
    \includegraphics[width=0.80\textwidth]{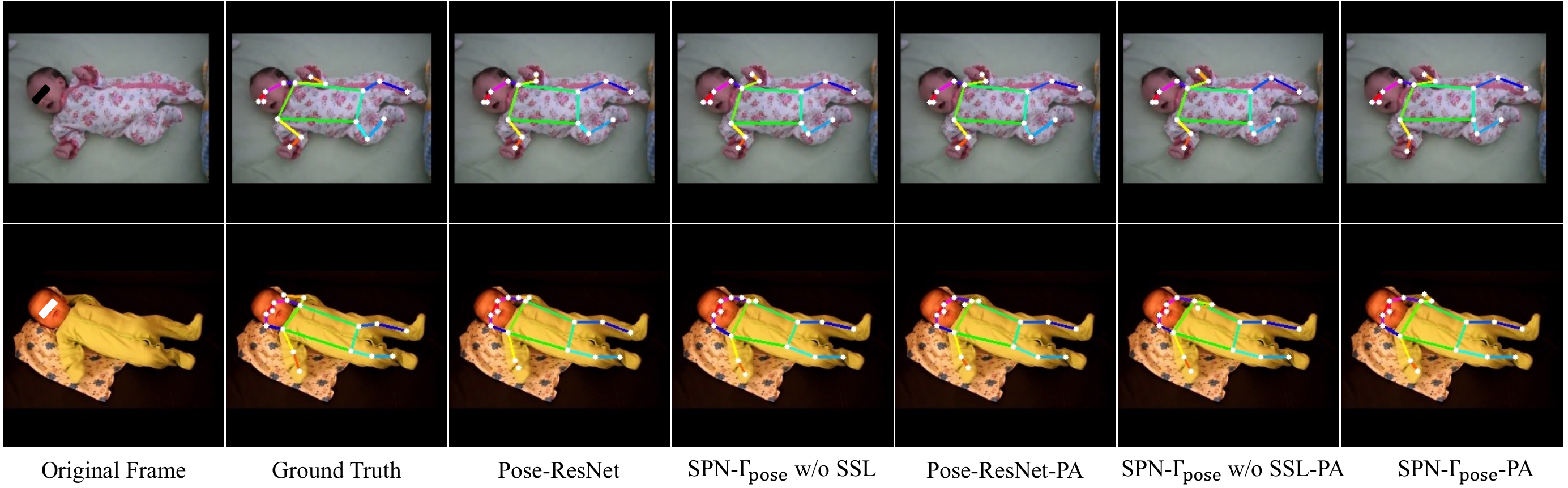}
    \caption{
    {Qualitative pose estimation comparison of SPN and Pose-ResNet on Youtube-Infant dataset. PA refers to models trained with partially labeled data.}
    }
    \label{fig:sota_pose}
\end{figure*}

To further demonstrate the power of semi-supervised learning in our proposed SPN, we experiment with a \textit{video-level} SSL setting on the BHT dataset: we randomly choose a certain number of training videos and remove \textbf{ALL} annotations from those videos. This setting is more stringent than the one used in \citep{jin2019incorporating}, which showed promising results with a \textit{frame-level} SSL setting: they removed labels of some frames in each video but all training video are still partially labeled. For our experiment using the \textit{video-level} SSL, we compare the performance of the [SPN w/o SSL] and the SPN with SSL when only keeping labeled frames in 7 training videos (and using 8 training videos without labels), and when only keeping labels in 5 videos (and using 10 without labels).

As shown in Table \ref{tab:rmv}, with fewer annotated videos, the SPN with SSL clearly shows significant performance gains (up to $6.31\%$) than [SPN w/o SSL] either using TAT or AAT with their best parameter setting ($\mathbb{T}=0.85$ and $t=0.4$). Such results indicate the potential of SPN under various semi-supervised learning scenarios. From both Table \ref{tab:semi} and Tabel \ref{tab:rmv}, one can find that the best model comes with using AAT ($t=0.4$), which we will set as the default SSL in the remaining experiments.

To validate the effectiveness of proposed FVGAN augmentation, we first utilize FVGAN to separately generate 10,000 images for each of the three types: synthesized foreground with target background $I_{\text{fore}}$, synthesized background with target foreground $I_{\text{back}}$, synthesized foreground and synthesized background $I_{\text{whole}}$. {Examples of synthesized frames can be found in Fig.~\ref{fig:fvgan_res2}. We then add all synthesized frames to the training sets for the training of [SPN w/o SSL] and [SPN w/ SSL], respectively. Note that, while there might be slight mismatches between the foreground and background lighting in the generated images, we believe, in the context of using these images for augmentation, the mismatch could further serve as a lighting augmentation to improve the robustness of the trained model.}


\begin{table}[t]
\centering
\caption{
{
Comparison between different methods of training data augmentation using different synthesized images on BHT dataset.}
}
\label{tab:fvgan10K}
\resizebox{0.7\linewidth}{!}{
\begin{tabular}{c|c|c|c|c}
\hline
                          & \multicolumn{2}{c|}{SPN w/o SSL} & \multicolumn{2}{c}{SPN w/ SSL}                              \\ \cline{2-5} 
\multirow{-2}{*}{Methods} & Dice             & Gain           & Dice                                  & Gain          \\ \hline
Baseline                  & 80.09            & 0              & 81.67                                 & 0             \\ \hline
10K $I_{\text{fore}}$                  & 83.29            & 3.20           & \textbf{84.99}                                 & \textbf{3.32}          \\
10K $I_{\text{back}}$                  & 82.08            & 1.99          & 82.51                                 & 0.84          \\
10K $I_{\text{whole}}$                 & \textbf{84.13}   & \textbf{4.04}  & {84.54} & {2.87} \\ \hline
\end{tabular}
}
\end{table}

Table \ref{tab:fvgan10K} shows our experimental results of using augmented training data with different generated images on BHT dataset. From Table \ref{tab:fvgan10K}, one can see that all three types of synthesized images can be helpful, 
and using fully synthesized $I_{\text{whole}}$ or foreground synthesized $I_{\text{fore}}$ shows better gain than background synthesized $I_{\text{back}}$, which can improve Dice coefficient by up to 4.04\% and 3.32\% for [SPN w/o SSL] and [SPN w/ SSL], respectively. We speculate the reason is that synthesizing foreground introduces more diverse infant body poses and appearances than synthesizing background, thus is more efficient for data augmentation.

Table \ref{tab:fvgan10K} also validates the effectiveness of our proposed semi-supervised learning.
Qualitatively, from Fig.~\ref{fig:ablation}, one can see that different from other variants of SPN, our full SPN model avoids mis-segmentation of shadows and occluded head region, giving the best segmentation performance.


\begin{table*}[ht]
\centering
\caption
{
Quantitative comparison between SPN and current state-of-the-art methods for infant body parsing on BHT dataset and Youtube-Infant dataset.
}
\label{tab:sota-both}
\resizebox{0.75\textwidth}{!}{%
\begin{tabular}{l|cccc|c|cccc|c}
\hline
\multirow{2}{*}{Methods}            & \multicolumn{5}{c|}{BHT}                                                      & \multicolumn{5}{c}{Youtube-Infant}                                           \\ \cline{2-11} 
                                    & Head           & Arm            & Torso          & Leg            & Dice           & Head           & Arm            & Torso          & Leg            & Dice           \\ \hline
U-Net \citep{ronneberger2015u}      & 37.27          & 29.81          & 45.42          & 60.22          & 47.16          & 56.34          & 31.52          & 60.79          & 34.54          & 46.95          \\
DeepLab V3+ \citep{chen2018encoder} & 72.00          & 53.93          & 69.99          & 73.29          & 70.99          & {91.76} & 72.18          & 86.26          & 82.87          & 84.16          \\
COSNet \citep{lu2019see}            & 76.68          & 60.50          & 82.17          & \textbf{80.38} & 79.09          & 90.02          & 72.37          & 84.70          & 82.10          & 83.02          \\
RVOS \citep{ventura2019rvos}        & \textbf{82.56} & {59.39} & \textbf{85.58} & 80.82          & \textbf{81.41} & 91.47          & \textbf{81.35} & \textbf{90.58} & \textbf{86.58} & \textbf{88.88} \\ 
Naive-Student \citep{chen2020naive} & 76.70          & \textbf{62.09}          & 77.86          & 79.55          & 77.10          & \textbf{92.51} & 73.49          & 86.24          & 82.34          & 84.30          \\ \hline
SPN w/o SSL                         & 75.34          & 62.26          & {81.35}        & 82.26          & 80.09          & 91.72          & 71.93          & 86.03          & 82.15          & 84.24          \\
SPN                                 & 74.38          & 64.33          & 83.18          & {83.51}        & {81.67}        & 92.51          & 75.64          & 87.68          & 85.22          & 86.48          \\
SPN w/o SSL + FVGAN                 & \textbf{82.36}        & {71.22} & \textbf{86.61}        & 84.43          & 84.13          & 93.34          & 77.92          & 88.46          & 86.78          & 87.19          \\
SPN + FVGAN                         & {82.25} & \textbf{71.77}        & {85.12} & \textbf{85.06} & \textbf{84.54} & \textbf{93.97} & \textbf{79.78} & \textbf{89.67} & \textbf{87.61} & \textbf{88.53} \\ \hline
\end{tabular}%
}

\end{table*}

\begin{table}[t]
\centering
\caption{{
Comparison of using different numbers of FVGAN synthesized $I_{\text{whole}}$ on BHT dataset and Youtube-Infant dataset for [SPN w/o SSL] method.}}
\label{tab:fvgan_ratio}
\resizebox{0.7\linewidth}{!}{%
\begin{tabular}{l|c|c|cc}
\hline
\multirow{2}{*}{Number} & \multicolumn{2}{c|}{BHT} & \multicolumn{2}{c}{Youtube-Infant} \\ \cline{2-5} 
                        & Dice           & Gain         & \multicolumn{1}{c|}{Dice}     & Gain    \\ \hline
0                       & 80.09          & 0            & \multicolumn{1}{c|}{84.24}    & 0       \\ \hline
2K $I_\text{whole}$                     & 82.62          & 2.53         & \multicolumn{1}{c|}{86.03}    & 1.79    \\
5K $I_\text{whole}$                    & 83.79          & 3.70         & \multicolumn{1}{c|}{86.64}    & 2.40    \\
10K $I_\text{whole}$                   & \textbf{84.13}          & \textbf{4.04}         & \multicolumn{1}{c|}{\textbf{87.19}}    & \textbf{2.95}    \\ \hline
\end{tabular}%
}
\end{table}

\begin{table}[ht]
\centering
\caption{{
Quantitative comparison between SPN and Pose-ResNet for pose estimation on Youtube-Infant dataset. PA refers to models trained with partially labeled data.
}
}
\resizebox{0.9\linewidth}{!}{%
\begin{tabular}{l|cc}
\hline
Methods               & PCK@0.1 & PCK@0.05 \\
\hline
Pose-ResNet~\citep{xiao2018simple}     & 99.04   & 95.11    \\
SPN-$\Gamma_{\text{pose}}$ w/o SSL     & \textbf{99.21}   & \textbf{96.69}   \\
\hline
Pose-ResNet-PA~\citep{xiao2018simple} & 97.78 & 94.09 \\
SPN-$\Gamma_{\text{pose}}$ w/o SSL-PA & 98.10   & 95.19    \\
SPN-$\Gamma_{\text{pose}}$-PA         & \textbf{99.12}   & \textbf{96.30}    \\
\hline
\end{tabular}%
}
\label{tab:spn-pose}
\end{table}

{
We further compare our proposed SPN and FVGAN with current state-of-the-art methods, including single frame based U-Net~\citep{ronneberger2015u} and DeepLab V3+~\citep{chen2018encoder}, and video based COSNet~\citep{lu2019see} and RVOS~\citep{ventura2019rvos}, and a semi-supervised video segmentation model Naive-Student~\citep{chen2020naive} in Fig.~\ref{fig:sota} and Table~\ref{tab:sota-both}.
}
For fair comparison, except U-Net, all methods employ pretrained models from ImageNet \citep{deng2009imagenet} or COCO dataset \citep{lin2014microsoft}. DenseCRF is also adopted as post-processing for all methods. 
From Fig.~\ref{fig:sota}, one can observe that SPN clearly handles occlusion better than other methods and shows better qualitative results.
As Table~\ref{tab:sota-both} shows, [SPN w/o SSL], and SPN, and [SPN w/o SSL + FVGAN], and the full [SPN + FVGAN] model all have achieved substantially better quantitative performance when compared with previous state-of-the-art methods {on BHT dataset}.

{To evaluate the generalizability of our proposed SPN and FVGAN, 
we also compare different SPN variants with current state-of-the-art methods on the Youtube-Infant dataset. As Table~\ref{tab:sota-both} and Fig.~\ref{fig:sota} show, SPN models achieve comparable or better performance when compared with state-of-the-art methods. 
Though RVOS \citep{ventura2019rvos} achieves the best mean Dice (0.35\% higher than [SPN + FVGAN]), note that the input of RVOS is a video clip while SPN only requires a pair of frames, thus SPN can be much more computationally efficient in training and testing.
The comparison between [SPN w/o SSL] and SPN (using AAT with $t=0.4$) in Table~\ref{tab:sota-both} and Fig.~\ref{fig:ablation} also demonstrates the effectiveness of our proposed semi-supervised learning strategy. Moreover, FVGAN augmented training improves the performance of both [SPN w/o SSL] and SPN, as demonstrated in Table~\ref{tab:sota-both} and Fig.~\ref{fig:ablation}.
We also experiment augmented training with different numbers of $I_\text{whole}$ images (2K, 5K, and 10K) for [SPN w/o SSL] method on BHT dataset and Youtube-Infant dataset. Table~\ref{tab:fvgan_ratio} shows the effectiveness of FVGAN augmented training using different numbers of synthesized frames, which can consistently improve the [SPN w/o SSL] by up to 4.04\% and 2.95\% in Dice score. Even augmenting with only 2K FVGAN synthesized frames can still boost the [SPN w/o SSL] model performance by $2.53\%$ and $1.79\%$ in Dice coefficient on the BHT and Youtube-Infant dataset, respectively.
}

{To verify that the improvement brought by FVGAN augmentation is statistically significant, we compare models trained with and without FVGAN using a one-sided paired Wilcoxon signed-rank test on the Youtube-Infant dataset. We calculate differences between two models from the mean Dice scores on each testing video. For models trained without SSL (\textit{i.e.}, comparing [SPN w/o SSL] with [SPN w/o SSL + FVGAN]), the p-value is $2.0\times10^{-5}$, and for models trained with SSL (\textit{i.e.}, comparing [SPN] with [SPN + FVGAN]), the p-value is $2.6\times10^{-5}$. Both p-values are considerably smaller than the considered significance level (\textit{i.e.}, 0.05), indicating that the improvement brought by FVGAN is indeed significant.}

%


{
Lastly, to better extract kinematic features for movement assessment, we extend our proposed SPN framework to the pose estimation task on Youtube-Infant dataset. Since Youtube-Infant is fully annotated for pose estimation, we first train a fully-supervised SPN and compare its pose estimation branch  $\Gamma_\text{pose}$ (\textit{i.e.}, the segmentation branch $\Gamma_\text{seg}$ in SPN for body parsing) with a strong baseline model Pose-ResNet \citep{xiao2018simple}. To evaluate the effect of our proposed semi-supervised learning mechanism, we remove 30\% labels of training frames to simulate partially annotated (PA) dataset and retrain different models as Pose-ResNet-PA, fully-supervised SPN-PA and semi-supervised SPN-PA. As shown in Table~\ref{tab:spn-pose}, when trained in a fully-supervised fashion, SPN can achieve 0.17\% and 1.58\% higher scores than Pose-ResNet on PCK@0.1 and PCK@0.05, respectively. Considering that the same backbone architecture is used between [SPN-$\Gamma_\text{pose}$] and Pose-ResNet, the major contribution to the improvement comes from the joint training of two branches in SPN. Table~\ref{tab:spn-pose} also illustrates the effectiveness of our proposed semi-supervised learning, which can improve [SPN w/o SSL] by 1.02\% and 1.11\% on PCK@0.1 and PCK@0.05, respectively. We also demonstrate some visualization results of SPN and Pose-ResNet in Fig.~\ref{fig:sota_pose}, from which one can observe that SPN models show better estimation in some joint keypoints such as the wrist.
}

\section{{Application to General Movement Assessment}}
\label{sec:gma}
{To further validate the clinical value of our proposed methods, we propose a convolutional recurrent neural network (CRNN) based CP prediction model for automatic GMA and conduct comprehensive experiments on a newly collected clinical IMV dataset to show that the body parsing and pose estimation results predicted by SPN can help improve CP prediction performance when combined with raw video frames.}

\subsection{{Dataset and Metrics}}
{We conduct comprehensive experiments on an IMV dataset collected from Shenzhen Baoan Women’s and Children’s Hospital (BWCH) with expert GMA annotations. The collection of the data was approved by the hospital ethics committee.
The BWCH dataset includes IMVs of 110 normal infants and 51 {abnormal} infants evaluated by an experienced pediatric physical therapist using the Prechtl's method of GMA \citep{ferrari2004prechtl}. GM videos were taken between 49 and 60 weeks. Each IMV is classified into {\it normal} or {\it abnormal} based on infant motion patterns. {Abnormal was defined as infants with larger amplitude movements than normal fidgety movements, such as with excessive speed and jerkiness~\citep{spittle2013general}. All videos have follow-up CP diagnoses that were verified to ensure the accuracy of the GMA annotation; in other words, all infants whose GM videos were annotated as abnormal in our dataset were later diagnosed as CP. }
We downsample the original videos every 5 frames and the average length of videos is about 900 frames.
We perform 5-fold cross validation on this dataset, and the ratio of normal to {abnormal} infants are kept at around 2:1 for both training and testing sets in each fold to preserve class distribution. We report the mean and variance of sensitivity, specificity, and AUC (Area Under the ROC curve) scores of all folds to compare different models.}

\subsection{{CRNN-based {GMA} Prediction}}
{We build upon a popular video classification framework CRNN \citep{shi2016end} to classify IMV into normal or {abnormal} as GMA prediction results. The original CRNN is based on raw video inputs, consisting of a CNN module to extract the feature map for each video frame and an LSTM model \citep{hochreiter1997long} taking as input the feature map sequence of all frames for final classification.
We enable CRNN to utilize body parsing or pose estimation predictions by adding an extra CNN to extract 2D feature maps from body parsing or pose estimation results of each frame. We subsequently concatenate these 2D feature maps with feature maps from raw video frames as inputs to LSTM.}

{To obtain better body parsing and pose estimation results for the testing IMVs in the new BWCH dataset, 
we retrain body parsing and pose estimation SPN models using all available data in the BHT dataset \citep{zhang2019online} and Youtube-Infant dataset \citep{chambers2020computer}. Then we directly apply these trained models to new testing IMVs without any training or fine-tuning. Since no video frame-level annotation is required, our setting simulates the real-world clinical scenario where no body parsing or pose estimation annotations are available for input IMVs.
For each input IMV, we first run the trained pose estimation model SPN-$\Gamma_{\text{pose}}$ to obtain the pose keypoint results. Based on the estimated keypoint coordinates of each frame, we generate a fixed bounding box to crop the video to remove the noisy background. We find that this preprocessing step is essential for achieving promising body parsing results. We then applied the trained body parsing SPN and trained SPN-$\Gamma_{\text{pose}}$ on the cropped video to generate the body parsing and pose estimation results for later use by GMA prediction. More implementation details are introduced in Section \ref{subsec:crnn_imp}.}

\begin{figure}[t]
    \centering
    \includegraphics[width=0.8\linewidth]{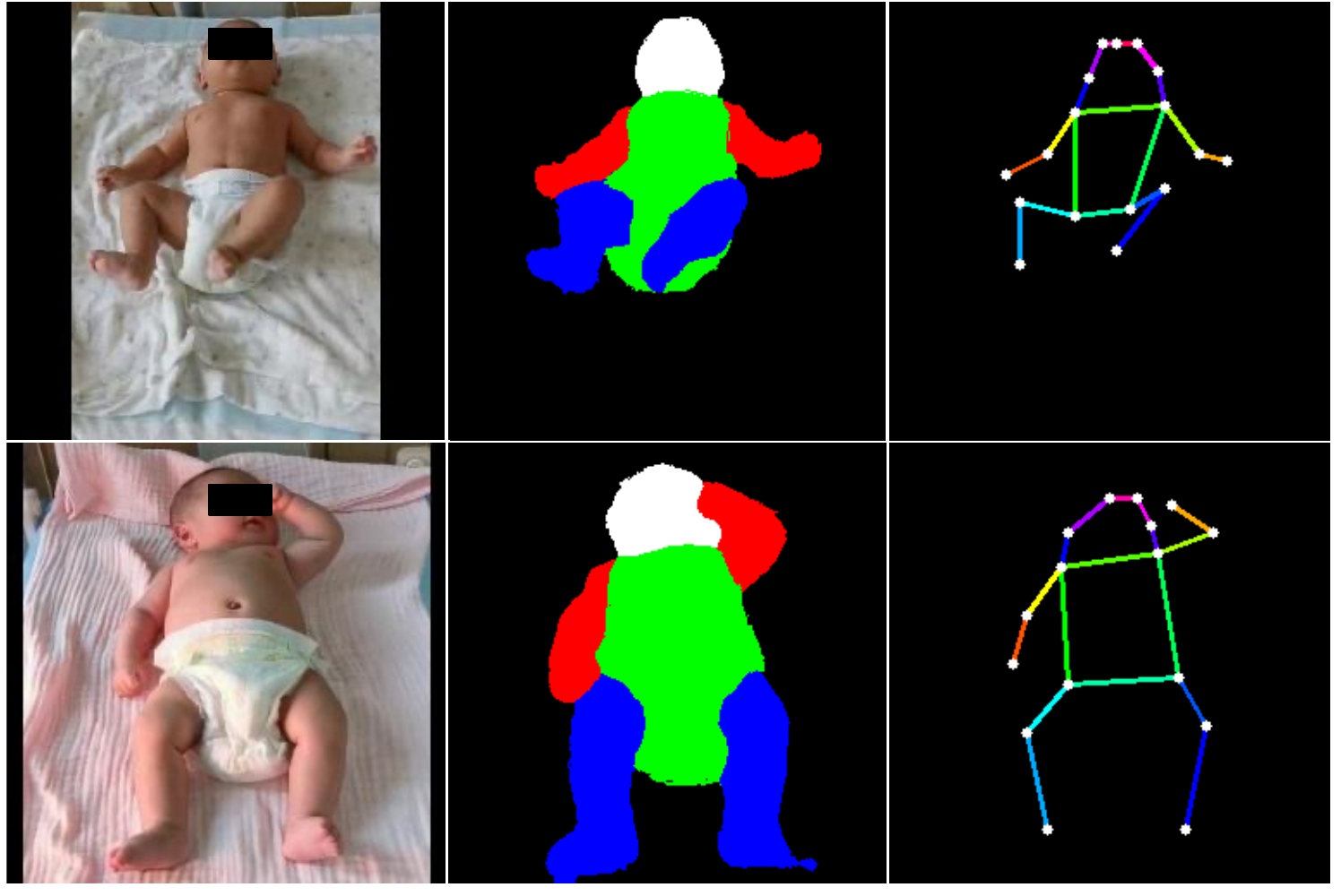}
    \caption{{Examples from BWCH dataset and corresponding body parsing and pose estimation results directly predicted by SPN without any fine-tuning.}}
    \label{fig:BWCH}
\end{figure}

\subsection{{Implementation Details}}
\label{subsec:crnn_imp}
{We adopt ResNet-152 \citep{he2016deep} and a three-layer LSTM to construct a CRNN model and initialize weights using a public model\footnote{\url{https://github.com/HHTseng/video-classification}} pretrained on the UCF101 \citep{soomro2012ucf101} action classification dataset. During the training, we fix the weights of ResNet-152 and fine-tune the remaining parts with a learning rate $1\times10^{-4}$ for 20 epochs in each cross-validation fold. The training batch size is set to be 32, and Adam \citep{kingma2014adam} optimizer is employed. Due to the long length of IMVs, we 
randomly select 200-continuous-frame video clips to train CRNN in each iteration. When testing, 
we clip the IMV into several 200-continuous-frame segments. The final prediction is obtained by taking an average of the class probability of each segment. To embed richer color information, we first represent body parsing and pose estimation results as RGB images (see Fig. \ref{fig:BWCH} as an example) and then use the pretrained ResNet-152 to extract their feature maps as supplement to the raw video input features.}

\subsection{{Result Analysis}}
{Table \ref{tab:gma} shows the quantitative comparison between the current state-of-the-art GMA prediction model \citep{chambers2020computer} and our proposed CRNN-based models using different input modalities. Within our proposed CRNN-based models, one can observe that using body parsing results (raw video + parsing), pose estimation predictions (raw video + pose), or their combination (raw video + pose + parsing) can all considerably improve the baseline CRNN model (raw video only). Such results further validate the clinical value of our proposed SPN. Furthermore, we find that body parsing features can generally obtain more gain than pose estimation features. The reason may be that body parsing representation contains richer information than the skeleton representation, thus is more appropriate for a CNN-based feature extraction model.
We also compare our proposed CRNN-based models with the state-of-the-art GMA method \citep{chambers2020computer} in Table \ref{tab:gma}. The original method in \cite{chambers2020computer} reported Bayesian Surprise concerning the reference population. For a fair comparison using the same cross-validation setting in our experiments, we first extract kinematic features for each IMV using their released implementation\footnote{\url{https://github.com/cchamber/Infant_movement_assessment}}. Then we train an XGBoost classifier \citep{chen2016xgboost} to evaluate extracted features on testing IMVs in each fold.
From Table \ref{tab:gma}, all CRNN models with SPN features achieve more favorable performance than \cite{chambers2020computer}, indicating that results generated by SPN can indeed help improve the GMA prediction performance under clinical settings.} 


\begin{table}[t]
\centering
\caption{{Quantitative comparison of different GMA methods on BWCH dataset.}}
\label{tab:gma}
\resizebox{0.95\linewidth}{!}{%
\begin{tabular}{l|cccc}
\hline
Methods              & Sensitivity & Specificity & AUC    \\
\hline
\cite{chambers2020computer}     & 48.69\textpm14.85    & 91.82\textpm1.82 & 70.25\textpm8.01 \\ \hline
CRNN (raw)           & 57.78\textpm25.33    & 92.73\textpm6.16 & 79.35\textpm15.97 \\
CRNN (raw + pose)    & 72.93\textpm20.17    & 95.45\textpm.07 & 94.02\textpm7.26 \\
CRNN (raw + parsing) & 80.61\textpm23.61    & \textbf{96.36}\textpm3.40 & 94.31\textpm8.31 \\
CRNN (raw + pose + parsing) & \textbf{82.43}\textpm15.63    & 87.27\textpm12.33 & \textbf{96.46}\textpm2.11 \\
\hline
\end{tabular}%
}
\end{table}

\section{Conclusion}
\label{sec:conclusion}
In this paper, we propose SiamParseNet, a novel semi-supervised framework for joint learning of body parsing and label propagation in IMVs toward computer-assisted GMA. Our proposed SPN exploits a large number of unlabeled frames in IMVs via alternative training of different training modes and shows superior performance under various semi-supervised training settings. Combined with factorized video GAN augmented training and multi-source inference for testing, SPN not only has great potential in infant body parsing 
{
but can also be easily adapted to other video tasks such as pose estimation. 
{A clinical GMA model for CP prediction can also benefit from the body parsing and pose estimation results generated by SPN.}
In the future, we plan to design a single SPN to jointly learn body parsing and pose estimation.}


\bibliographystyle{model2-names.bst}\biboptions{authoryear}
\bibliography{mybib.bib}



\end{document}